\documentclass{article}
\usepackage{colm2024_conference}

\usepackage{booktabs}
\usepackage{graphicx}
\usepackage{enumitem}
\usepackage{wrapfig}
\usepackage{natbib}
\usepackage{makecell}
\usepackage{array}
\usepackage{amsmath}
\usepackage{amsfonts}
\usepackage{multirow}
\usepackage{verbatim}
\usepackage{caption}
\usepackage{longtable}
\usepackage{ragged2e}
\usepackage{array}
\usepackage[utf8]{inputenc}
\usepackage[T1]{fontenc}
\usepackage{pifont}
\usepackage{xcolor}
\usepackage{tablefootnote}
\usepackage{xspace}
\usepackage{textcomp}
\usepackage{lscape}
\usepackage{siunitx}
\usepackage{listings}
\usepackage{colortbl}
\hypersetup{colorlinks=true}
\usepackage{url}
\usepackage{microtype}
\usepackage{subcaption}
\usepackage{fancyvrb}
\usepackage{fvextra}
\definecolor{graybg}{gray}{0.95}
\definecolor{subcol}{HTML}{EBF5FB} 
\definecolor{catgray}{HTML}{F0F0F0}

\newcommand{\cmark}{\ding{51}}

\usepackage[capitalize,noabbrev]{cleveref}
\crefname{section}{Section}{\S\S}
\Crefname{section}{Section}{\S\S}
\crefname{table}{Table}{Tables}
\crefname{figure}{Figure}{Figures}
\crefname{equation}{eq.}{}
\crefname{appendix}{Appendix}{}
\crefformat{section}{Section #2#1#3}

\usepackage{titlesec}
\titleformat*{\section}{\large\bfseries}

\title{Video-MME-v2: Towards the Next Stage in Benchmarks \\ for Comprehensive Video Understanding}

\author{
Video-MME Team\thanks{Full authors are listed in contributions.}
\\
\footnotesize{
\url{https://video-mme-v2.netlify.app/}
}
}

\begin{document}

\maketitle

\begin{abstract}
With the rapid advancement of video understanding, existing benchmarks are becoming increasingly saturated, exposing a critical discrepancy between inflated leaderboard scores and real-world model capabilities. To address this widening gap, we introduce \textbf{Video-MME-v2}, a comprehensive benchmark designed to rigorously evaluate the robustness and faithfulness of video understanding. To systematically evaluate model capabilities, we design a \textbf{progressive tri-level hierarchy} that incrementally increases the complexity of video comprehension, ranging from multi-point visual information aggregation, to temporal dynamics modeling, and ultimately to complex multimodal reasoning. Besides, in contrast to conventional per-question accuracy, we propose a \textbf{group-based non-linear evaluation} strategy that enforces both consistency across related queries and coherence in multi-step reasoning. It penalizes fragmented or guess-based correctness and assigns credit only to answers supported by valid reasoning. To guarantee data quality, Video-MME-v2 is constructed through a rigorously controlled human annotation pipeline, involving 12 annotators and 50 independent reviewers. Backed by \textbf{3,300 human-hours} and up to \textbf{5 rounds} of quality assurance, Video-MME-v2 aims to serve as one of the most authoritative video benchmarks. Extensive experiments reveal a substantial gap between current best model Gemini-3-Pro and human experts (\textbf{49.4 vs. 90.7}), and uncover a clear hierarchical bottleneck where errors in visual information aggregation and temporal modeling propagate to limit high-level reasoning. We further find that thinking-based reasoning is highly dependent on textual cues, improving performance with subtitles but sometimes degrading it in purely visual settings. By exposing these limitations, Video-MME-v2 establishes a demanding new testbed for the development of next-generation video MLLMs.
\end{abstract}

\section{Introduction}

Recent advancements~\cite{zhang2024llava,zhang2025videollama,bai2025qwen25vltechnicalreport,bai2025qwen3,feng2025videor1reinforcingvideoreasoning,li2025videochatr1enhancingspatiotemporalperception,gemini3pro} in video-based multimodal large language models (video MLLMs) have led to remarkable progress across a variety of understanding and reasoning tasks, signaling a shift from simple video comprehension to deeper understanding and more complex reasoning. 
Despite this progress, existing evaluations~\cite{fu2025videommefirstevercomprehensiveevaluation,wu2024longvideobenchbenchmarklongcontextinterleaved,li2024mvbenchcomprehensivemultimodalvideo,hong2025motionbenchbenchmarkingimprovingfinegrained} often lack a comprehensive evaluation hierarchy, emphasizing performance on task-specific benchmarks or isolated topics, which makes it difficult for holistic assessment. Furthermore, previous work mainly focuses on per-question accuracies, overlooking the need for consistent and trustworthy video comprehension in evaluation. These limitations hinder a thorough assessment of frontier video MLLMs and underscore the need for deeper investigation into their robust understanding and reliable reasoning capabilities.

To address these challenges, we introduce Video-MME-v2, a comprehensive benchmark designed to evaluate the robustness and faithfulness of video MLLMs for dynamic visual content comprehension. This is achieved through a novel \textbf{multi-level evaluation hierarchy} and a \textbf{group-based evaluation strategy}.
\begin{itemize}
    \item \textbf{Multi-level Evaluation Hierarchy.} Previous general video understanding benchmarks often treat different capabilities as separate aspects, lacking a comprehensive taxonomy for holistic evaluation. Our evaluation hierarchy categorizes core video understanding skills into three progressive levels. Level 1 focuses on information aggregation, assessing the model's ability to perceive and aggregate cross-frame and cross-modal information. Level 2 examines temporal dynamics modeling, evaluating the capture of causality, state changes, and sequential order. Level 3 targets complex video reasoning, mimicking real-world scenarios to test advanced video comprehension skills such as physical understanding, social intelligence, and complex plot comprehension. Together, this progressive framework, from multiple-point information aggregation to temporal dynamics modeling, and finally to complex reasoning, ensures a holistic evaluation from foundational visual perception to sophisticated, human-like video comprehension.
    \item \textbf{Group-based Evaluation Strategy.} Our group-based evaluation strategy assesses frontier models from two different perspectives: (1) \textbf{Capability consistency}, which examines the breadth of a specific fundamental perception skill through groups of tasks varying in aspect (e.g., from relative position judgment to spatial reconstruction for spatial understanding) and granularity (e.g., from object counting to motion counting for video-based counting); and (2) \textbf{Reasoning coherence}, which measures the depth of a model’s reasoning ability by presenting sequences of temporally and causally related questions that reveal whether the model can follow logical steps toward complex, high-level inference. To synergize with this group-based design, we further introduce a non-linear scoring method that evaluates the joint correctness of correlated questions rather than treating them independently. It penalizes fragmented or guess-based success and enforces stepwise reasoning validity. By jointly evaluating the breadth of foundational skills and the depth of complex reasoning, this strategy provides a rigorous measure of a model's capability of genuine multimodal video proficiency.
\end{itemize}
Complementing our benchmark design, we develop a meticulous human annotation pipeline with substantial human involvement to ensure high data quality. In total, \textbf{12 annotators and 50 reviewers} contribute more than \textbf{3,300 human-hours}. We enforce strict video selection criteria and implement a rigorous multi-stage review and quality control process. Following this pipeline, we curate a dataset of \textbf{800 videos and 3,200 questions}.
Together, Video-MME-v2 provides a high-quality and comprehensive benchmark that assesses not only isolated task performance but also a model's ability to achieve robust, faithful comprehension in complex video scenarios.

To demonstrate the challenging nature of Video-MME-v2, we conduct extensive evaluations across a broad spectrum of frontier video MLLMs. Our analysis encompasses cutting-edge proprietary systems, such as Gemini-3-Pro~\cite{gemini3pro}, GPT-5~\cite{gpt5}, Seed-2.0~\cite{bytedance2026seed2modelcard}, and MiMo-v2-Omni~\cite{xiaomi_mimo_v2_omni}, alongside leading open-source models like Qwen3.5~\cite{qwen3.5} and Kimi-K2.5~\cite{team2026kimi}. 
The empirical results reveal a substantial gap between current models and human experts: while human experts achieve a score of 90.7, the best-performing model, Gemini-3-Pro, reaches only 49.4. A significant performance gap also persists in the open-source community, where the top-performing model, Qwen3.5-397B-A17B-Think, achieves 39.1. 
Beyond this gap, we uncover a clear hierarchical bottleneck in video understanding. Failures in high-level reasoning are not solely due to insufficient reasoning ability, but are also caused by errors accumulated in earlier stages, including visual information aggregation and temporal modeling.
These results expose a critical limitation: while current models may perform adequately on shallow, perception-level tasks, \textbf{they fundamentally lack the capability consistency and reasoning coherence required to navigate dynamic, real-world scenarios}. 

We further conduct analysis experiments on Video-MME-v2 and yield several key insights. 1) Conventional per‑question accuracy substantially overestimates model capability, whereas our group‑based nonlinear scoring reveals that even state‑of‑the‑art models lack consistency across correlated queries. 2) Interestingly, enabling thinking modes improves performance with subtitles but can cause significant regression without textual cues, indicating current models still overweight language-based reasoning, resulting in an over‑reliance on language priors. 3) Omni-modal aggregation, long-context temporal modeling, and complex reasoning demonstrate synergistic improvement, yet large parameter scales can partially compensate for missing capabilities. To this end, we hope that the Video-MME-v2 benchmark, supported by these experiments and in-depth analyses, will serve as a flagship standard for video MLLM evaluation and drive future development in the field.
\section{Related Work}

\paragraph{Advancements in Video MLLMs.}
Video understanding is a crucial research direction for multimodal large language models (MLLMs). Early studies such as LLaVA-Video~\cite{zhang2024llava} and Qwen2.5-VL~\cite{bai2025qwen25vltechnicalreport} primarily treat videos as sequences of individual frames, thereby transferring capabilities from image understanding to video comprehension. With the rapid progress of MLLMs in visual reasoning, recent works~\cite{feng2025videor1reinforcingvideoreasoning,li2025videochatr1enhancingspatiotemporalperception,yan2025videochatr15visualtesttimescaling,dong2026icl,tian2025egor1chainoftoolthoughtultralongegocentric,dong2026insight} have begun to tackle more complex video reasoning tasks. By incorporating Group Relative Policy Optimization (GRPO), models like Video-R1~\cite{feng2025videor1reinforcingvideoreasoning} and VideoChat-R1~\cite{li2025videochatr1enhancingspatiotemporalperception} have shown enhanced reasoning abilities for video tasks. More recent developments, such as VideoChat-R1.5~\cite{yan2025videochatr15visualtesttimescaling}, further extend this line of research by integrating tool usage into video understanding.

\paragraph{Video MLLMs Benchmarks.}
With the rapid advancement of video MLLMs, corresponding benchmarks have also evolved. Existing efforts mainly fall into two categories. Some focus on domain-specific capabilities, such as MVBench~\cite{li2024mvbenchcomprehensivemultimodalvideo} and MotionBench~\cite{hong2025motionbenchbenchmarkingimprovingfinegrained}, which emphasize fine-grained action understanding, and LongVideoBench~\cite{wu2024longvideobenchbenchmarklongcontextinterleaved} and LVBench~\cite{wang2025lvbenchextremelongvideo}, which target long video comprehension. Meanwhile, benchmarks such as Video-MME~\cite{fu2025videommefirstevercomprehensiveevaluation} aim to provide a more comprehensive yet relatively basic evaluation of general video understanding abilities. More recent works, including VideoMMMU~\cite{hu2025videommmuevaluatingknowledgeacquisition}, MMVU~\cite{zhao2025mmvumeasuringexpertlevelmultidiscipline}, and VideoReasonBench~\cite{liu2025videoreasonbenchmllmsperformvisioncentric}, have shifted attention toward complex video reasoning, marking an emerging trend in video MLLM research. However, existing benchmarks tend to evaluate either specific or elementary capabilities, leaving the deeper investigation on perception and reasoning largely unexplored. To address this gap, we introduce Video-MME-v2, a new benchmark designed to assess both perception consistency and reasoning coherence through a comprehensive and multi-level task suite, serving as a robust evaluation standard for the next generation of video MLLMs.
\section{Benchmark Design}

\subsection{Progressive Capability Hierarchy}

\begin{figure}[t]
  \centering
  \includegraphics[width=\linewidth]{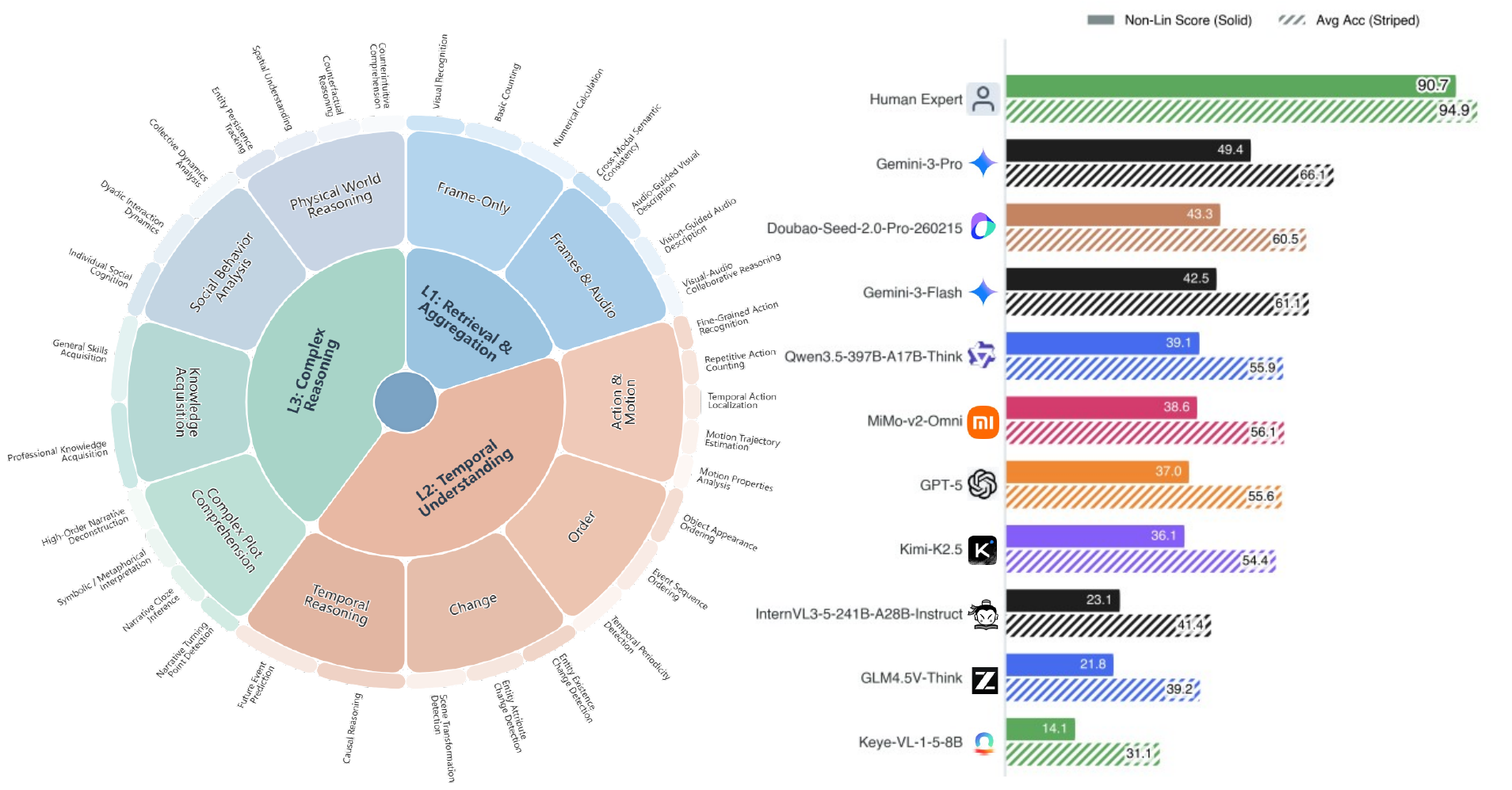}
\caption{\textbf{Left}: The three-level capability hierarchy of Video-MME-v2: distribution of capability dimensions across Level 1 (information retrieval and aggregation), Level 2 (temporal understanding), and Level 3 (complex reasoning). \textbf{Right}: Models are ranked by their group-based non-linear scores, while average accuracy is provided for reference only. Due to API limitations, Gemini models are tested by extracting and compressing video frames to 60M, while GPT-5 is tested with an input of 50 frames.}  \label{fig:teaser}
\end{figure}

Previous benchmarks predominantly construct questions from isolated aspects, omitting the progressive nature of video comprehension. To address this limitation, we organize our dataset into three hierarchical levels comprising 12 sub‑categories and over 30 task types, spanning from basic multiple‑point information aggregation through dynamic temporal modeling to complex reasoning. As shown in Figure~\ref{fig:teaser}, questions are evenly distributed across these levels. A complete taxonomy is provided in Table~\ref{tab:taxonomy_full}.

\begin{itemize}[leftmargin=*, noitemsep]
  \renewcommand\labelitemi{$\diamond$}   
    \item \textbf{Level 1: Visual Information Aggregation.} 
    This foundational level primarily assesses the model's ability to identify and integrate information at specific timestamps. It encompasses three key aspects: \textit{Visual Recognition}, which focuses on identifying objects, attributes, and scenes; \textit{Cross-Modal Consistency}, which evaluates the alignment of visual cues with audio semantics (e.g., tone-mood synchronization); and \textit{Basic Counting \& Calculation}, involving tasks related to counting and basic numerical reasoning.

    \item \textbf{Level 2: Temporal Dynamics.} 
    Building upon Level 1, this level emphasizes the temporal evolution of events. It includes three main sublevels: \textit{Action \& Motion Analysis}, which assesses the model’s ability to recognize actions and track motion trajectories; \textit{Sequential Ordering}, which focuses on determining the chronological order of events or object appearances; and \textit{Causal Reasoning}, which involves understanding the cause of events based on their temporal context.

    \item \textbf{Level 3: Complex Reasoning.} 
    This advanced level simulates real-world cognitive tasks requiring professional knowledge and multi-hop inference. It contains three major aspects: \textit{Narrative Understanding}, which involves interpreting plot twists, metaphors, and non-linear storytelling; \textit{Social Dynamics}, which focuses on analyzing dyadic interactions and collective behaviors; and \textit{Physical World Reasoning}, which requires counterfactual reasoning and understanding physical constraints in real-world scenarios.
\end{itemize}

\subsection{Group Type Definition}
\label{sec:group}
Previous work has primarily focused on per-question accuracy, overlooking the importance of consistent and coherent comprehension for faithful video understanding and reasoning. In the image domain, MME~\cite{fu2025mmecomprehensiveevaluationbenchmark} first introduces augmented group QA (a target question can be asked in two different ways, and the answers are respectively `yes' and `no') to test understanding reliability. MMBench~\cite{liu2024mmbenchmultimodalmodelallaround} proposes a circular evaluation strategy that swaps answer option content to ensure answer stability. In video understanding, Video-TT~\cite{zhang2025videothinkingtestholistic} is the first benchmark to employ augmented group questions to probe model consistency. However, these methods largely concentrate on augmenting individual questions, while neglecting the significance of interrelations within a group. Our design extends this perspective by incorporating question groups that explicitly model the relationships among related queries for both perception and reasoning, enabling a more comprehensive assessment of model understanding.

\paragraph{Consistency-Based Group.}
The consistency-based group aims to address a key problem: how to evaluate a specific capability using a limited set of questions, thereby enabling a more accurate assessment of model competence. We construct consistency groups along two dimensions: breadth and granularity.
For breadth, we design diverse question types within a single domain to capture different reasoning aspects. For example, within the spatial understanding domain, we include object localization consistency and relative motion reasoning questions to evaluate both static and dynamic spatial comprehension.
For granularity, we extend one question type across multiple spatio-temporal scales. For instance, in one fitness tutorial video, we ask about the global sequence of exercises to assess holistic understanding, and the ordering of sub-actions within a single move to measure fine-grained action comprehension.
By combining these two dimensions, the consistency-based groups provide a systematic and multi-granularity evaluation of video understanding within a specific domain.

\paragraph{Coherence-Based Group.}
The coherence-based group is designed to evaluate a model’s reasoning coherence in handling complex video reasoning tasks. Existing benchmarks~\cite{hu2025videommmuevaluatingknowledgeacquisition,zhao2025mmvumeasuringexpertlevelmultidiscipline} typically assess only the final answers to such questions, overlooking how the model arrives at them. However, tracing the intermediate reasoning process is crucial for determining whether the model genuinely performs multi-step reasoning or merely guesses the correct option by chance. While some works~\cite{jiang2025mme,qi2025vcrbenchcomprehensiveevaluationframework} have explored intermediate supervision by evaluating reasoning content, they do not structure the evaluation as a progressive question group that explicitly checks milestone reasoning steps from the outset. In each coherence-based group, our question sets are constructed to mimic the logical progression a human would follow to solve a complex problem. For instance, for a complex plot analysis task, where a character fakes their death to deceive others, we systematically evaluate whether the model can identify direct visual clues of the apparent death, capture anomalous details that contradict a genuine death, infer the underlying purpose of the staged act, and draw a final conclusion constrained by the preceding evidence. This hierarchical verification process through clue localization, anomaly verification, purpose explanation, and conclusion closure allows us to efficiently distinguish whether the model conducts coherent reasoning grounded in the video content. This design enables us to establish an explicit reasoning chain, providing a more rigorous and interpretable assessment of the true reasoning capability of current video MLLMs.

\subsection{Metrics}
Our evaluation incorporates both conventional per-question accuracy and group-level scores. To align with the fundamental design philosophy of Video-MME-v2, we apply a non-linear scoring mechanism to the group evaluations, explicitly emphasizing capability consistency and reasoning coherence.

\paragraph{Average Accuracy (Avg Acc).}
Let $q$ index questions. Avg Acc is the mean of per-question correctness:
\begin{equation}
\mathrm{AvgAcc} = \frac{1}{|\mathcal{Q}|}\sum_{q \in \mathcal{Q}} \mathbb{I}[\hat{y}_q = y_q].
\end{equation}

\paragraph{Group-level Non-linear Score.} Conventional evaluation protocols rarely employ non-linear scoring metrics. In the image domain, MME~\cite{fu2025mmecomprehensiveevaluationbenchmark} pioneered a non-linear approach by awarding bonus points to models that correctly answer augmented questions tied to the same task. However, in the video domain, prior works predominantly rely on standard average accuracy, largely overlooking how careful metric design can better reflect the robustness and faithfulness of video comprehension. In Video-MME-v2, we introduce a group-level non-linear metric designed to evaluate a model's robustness against related questions within a specific group. 
Let $g$ index groups, each group has four questions $(q_{g,1},\dots,q_{g,4})$. We define a group score $S(g)$ that depends on the \emph{joint} correctness pattern within a group, rather than treating questions independently:
\begin{equation}
\mathrm{Overall} = \frac{1}{|\mathcal{G}|}\sum_{g \in \mathcal{G}} S(g).
\end{equation}
For consistency groups, we use a non-linear scoring function: given $\mathcal{N}$ correct answers out of $4$ related questions, the group score is $(\mathcal{N}/4)^2$. This quadratic suppression penalizes isolated correct guesses and rewards consistent performance across different facets of the same capability. For coherence groups, we apply a \emph{first-error truncation} mechanism: starting from the first reasoning step, only the longest consecutive sequence of correct answers counts toward the score. Once an error occurs, any later correct answers are ignored. This prevents the model from earning credit for correct but logically unsupported steps, ensuring that only faithful reasoning chains are rewarded.

\section{Dataset Construction and Annotation}

The evaluation system of Video-MME-v2 places extremely high demands on annotation quality. To this end, we establish a comprehensive and rigorous data annotation and quality control process. After investing \textbf{3,300 human-hours of 12 annotators and 50 reviewers}, we ultimately collect 800 videos, each paired with 4 questions and 8 answer options per question. The pipeline is designed to (i) minimize potential pretraining leakage via recency-oriented curation and explicit decontamination, (ii) ensure broad real-world coverage via a hierarchical category taxonomy, and (iii) enforce strong multimodal dependence and high discriminability via group-based question design, adversarial option construction, and rigorous verification procedures.

\subsection{Video Curation}
\label{subsec:video_curation}

\paragraph{Recency-Oriented Collection To Mitigate Leakage.}
To fundamentally mitigate the risk of data contamination and ensure the benchmark's timeliness, we source videos from the internet with a strict focus on \emph{temporal recency}. The collection process specifically targeted newly released content. Over 80\% of the videos in our dataset are published in 2025 or later, with nearly 40\% published after October 2025. This temporal boundary ensures they are highly unlikely to have been included in the pre-training corpora of current MLLMs, thus reflecting the models' true reasoning capabilities rather than mere memorization.

\begin{figure}[t]
    \centering
    \begin{minipage}{0.49\linewidth}
        \centering
        \includegraphics[width=\linewidth]{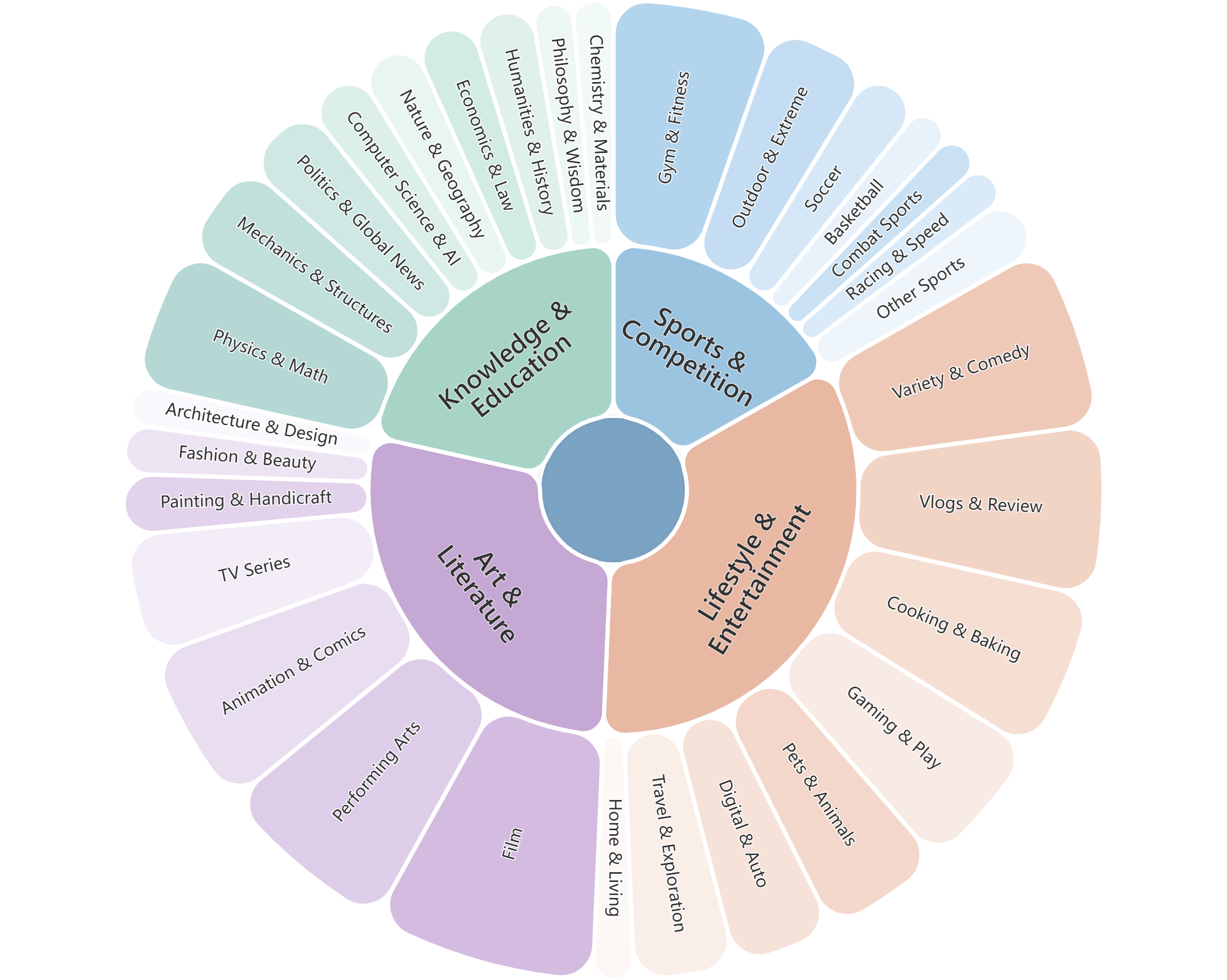}
        \captionof{figure}{Video category distribution.}
        \label{fig:category}
    \end{minipage}
    \begin{minipage}{0.49\linewidth}
        \centering
        \includegraphics[width=\linewidth]{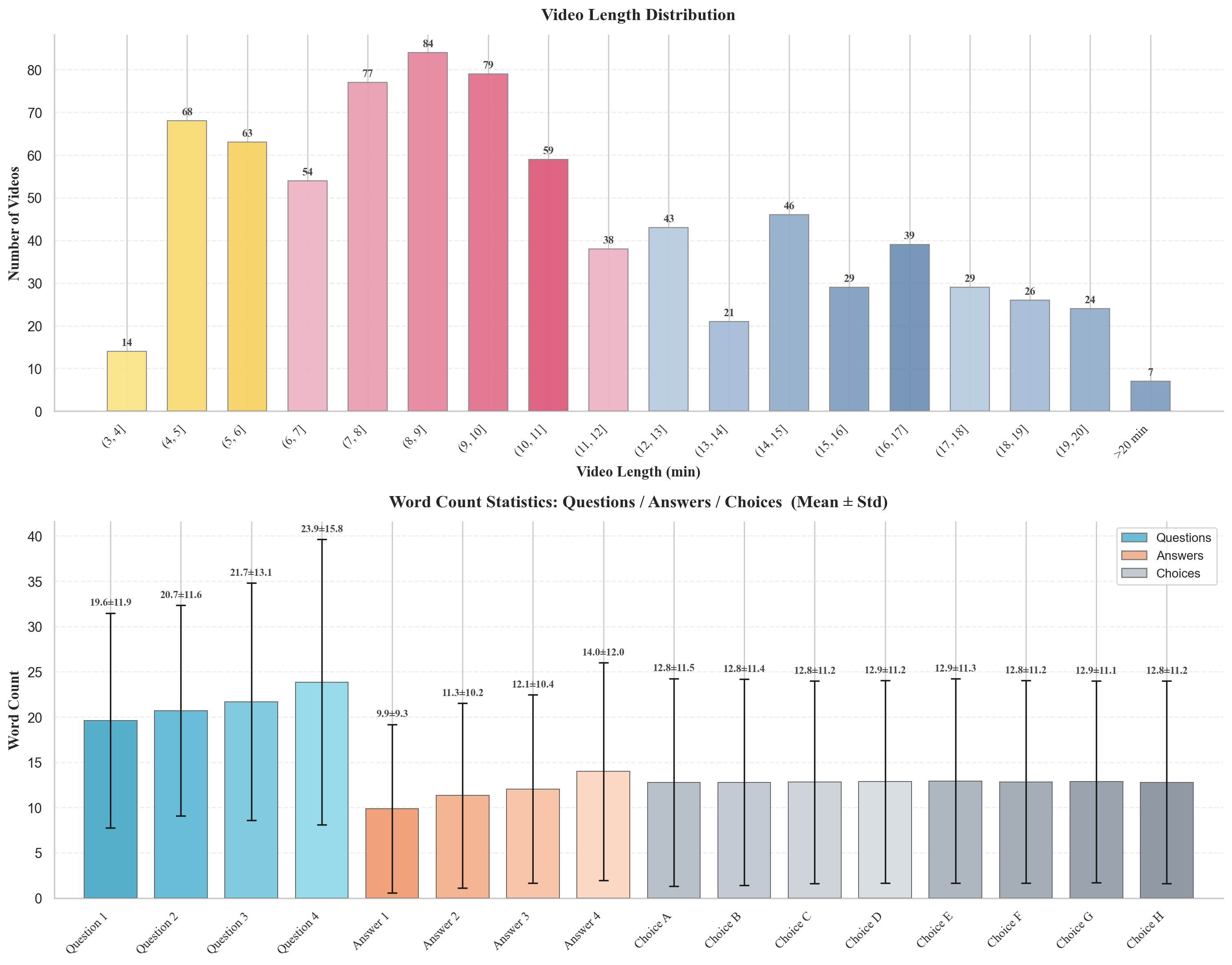}
        \captionof{figure}{Video length and word count statistics.}
        \label{fig:data_stats}
    \end{minipage}

    \vspace{0.8em}

    \begin{minipage}{0.49\linewidth}
        \centering
        \includegraphics[width=\linewidth]{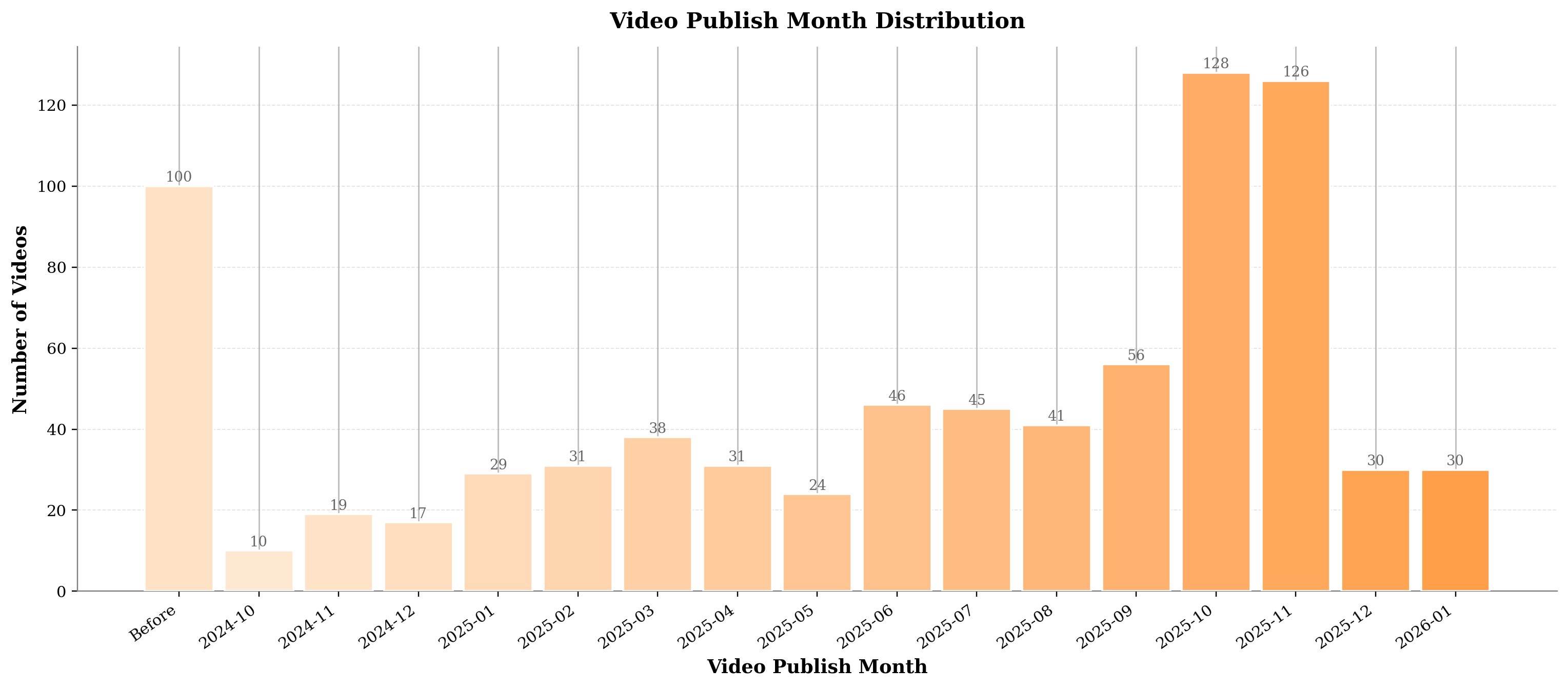}
        \captionof{figure}{Video release-time distribution.}
        \label{fig:publish_month}
    \end{minipage}\hfill
    \begin{minipage}{0.49\linewidth}
        \centering
        \includegraphics[width=\linewidth]{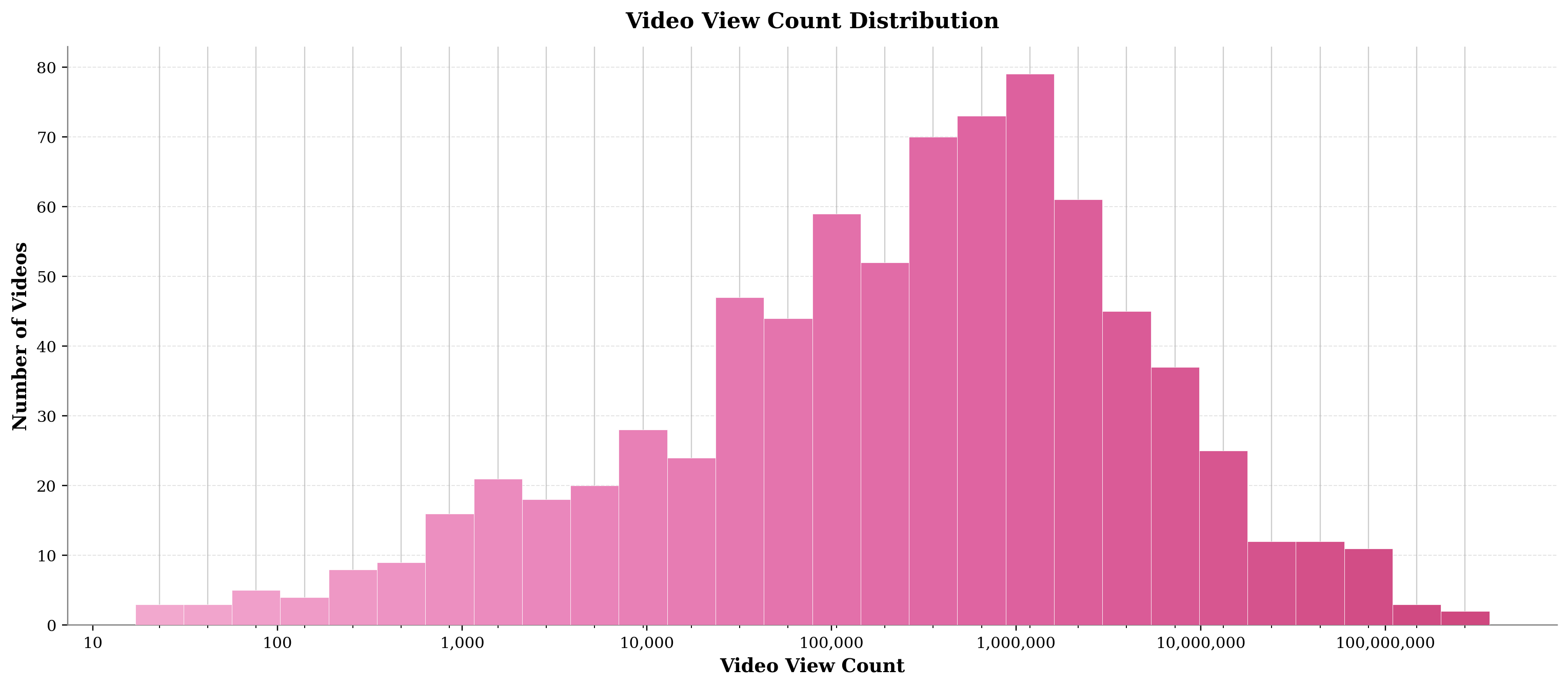}
        \captionof{figure}{Video view-count distribution.}
        \label{fig:viewcount}
    \end{minipage}
\end{figure}

\paragraph{Diversity via A Hierarchical Taxonomy.}
To reflect realistic deployment scenarios, we organize sources into a two-level taxonomy with four top-level domains: \textit{Sports \& Competition}, \textit{Lifestyle \& Entertainment}, \textit{Art \& Literature}, and \textit{Knowledge \& Education}. These are further split into 31 fine-grained subcategories (e.g., cooking, travel, film, physics, computer science). \Cref{fig:category} shows the category distribution, which is intentionally balanced and diversified across both subject matter and visual styles. Furthermore, as shown in \Cref{fig:data_stats} (top), the videos span an average duration of 10.4 minutes, with 99\% under 20 minutes and 53\% under 10 minutes, forming a diverse distribution tailored for comprehensive evaluation.

\paragraph{Quality Control with View-Count Thresholding.}
To minimize low-quality and noisy samples at the source, we use view counts as a proxy for content quality. As shown in \Cref{fig:viewcount}, the mean and median view counts are 4.83 million and 355 thousand respectively. We filter out videos with low exposure; 84.3\% of the selected videos have exceeded 10,000 views, and 94.4\% exceed 1,000 views, ensuring that the benchmark consists of high-quality, human-curated content.

\paragraph{Manual Decontamination.}
To further shield the evaluation from the ``memorization effect,'' we perform manual screening to exclude classic films, television works, and flagship content from top-tier influencers. This step ensures that model performance reflects genuine perception and reasoning rather than the retrieval of training-time memories.

\paragraph{Length Distribution of Questions and Options.}
As shown in \Cref{fig:data_stats} (bottom), the average length of questions and answers exhibits a progressive increase from Q1 to Q4. This structural trend aligns with our \textit{Reasoning Coherence} design philosophy: questions placed later in the sequence inherently demand greater logical depth, necessitating more comprehensive contextual descriptions and more detailed, nuanced answers. Meanwhile, the average word count across the eight options remains highly consistent, ensuring that models cannot exploit length bias to guess the correct answer.

\subsection{Question and Option Design}
\label{subsec:question_design}

\paragraph{Group-Based Construction.}
As previously mentioned, we adopt a group-based construction to ensure both \textit{capability consistency} (breadth) and \textit{reasoning coherence} (depth). As shown in the statistical trends of question lengths (\Cref{fig:data_stats}), the average length of questions and answers progressively increases from Q1 to Q4. This deliberate structural design inherently creates logical chains that challenge the model’s fine-grained understanding, where later questions are more difficult and typically require more sufficient contextual descriptions and more detailed answers.

\paragraph{Rigorous Verification with Frontier Models.}
Throughout the annotation process, we conduct real-time validation and adversarial stress-testing using frontier models (e.g., Gemini-3-Pro). This cross-check ensures the precision of question phrasing and the rigor of ground-truth settings. Candidate questions are continuously tested to detect underspecified premises, language-only shortcuts, and weak distractors, and are iteratively revised until they exhibit sufficient discriminability and multimodal dependence.

\paragraph{Low-Guessing Baseline with High-Confusion Options.}
To maximize evaluation resolution and discourage superficial pattern matching, each question features an 8-option multiple-choice design (A--H), reducing the random-guessing probability to just 12.5\%. Furthermore, as depicted in \Cref{fig:data_stats} (bottom), the average word count across the eight options is strictly controlled to be highly consistent, effectively eliminating any length-based biases or statistical shortcuts.

\paragraph{Strong Distractor Design.}
Beyond multiple conventional strong distractors, we require at least one \emph{adversarial distractor} per question—a meticulously crafted option that is highly plausible given partial visual or audio evidence yet directly contradicts a key, fine-grained detail in the ground truth. Distractors are initially drafted from model-generated candidates and then undergo fine-grained manual refinement by annotators to precisely test the model's discriminative capabilities without leaking the correct answer.
\subsection{Quality Assurance}
\label{subsec:qa}

We recruit an independent quality-control team of 50 independent reviewers to minimize subjective bias from the 12 human annotators and the influence of large model priors on data quality. Each video sample is independently reviewed by at least 2 quality-control personnel. The team then collaborate with the annotation team to review and confirm all revisions, completing all dataset questions themselves to calculate human accuracy rates.

\paragraph{Text-Only Baseline Testing.} 
We use {Gemini-3-Pro} in a text-only mode as a baseline to identify and remove questions that can be solved without visual information. This step strictly controls language priors and ensures the necessity of multimodal perception for answering the questions correctly.

\paragraph{Multi-Round Cross-Validation.} 
Every question undergoes three rounds of cross-review by different annotators. This intensive peer-review process aims to eliminate semantic ambiguity, patch potential logic loopholes, and fine-tune the design of the strong distractors.

\paragraph{Independent Blind Testing.} 
The 50 independent evaluators, who are not involved in the original annotation phase, perform multiple rounds of fine-grained checks on each video-question pair. This external validation significantly reduces subjective bias and ensures the broader generalizability and clarity of the benchmark.

\paragraph{Closed-Loop Iterative Verification.} 
We establish a ``Correction-Reverification'' mechanism. Any modified question is re-subjected to the text-only baseline test and independent blind testing. This closed loop ensures that every revision leads to a controllable and verifiable improvement in quality, maintaining the integrity of the dataset.

\section{Experiments}

We evaluate a diverse array of state-of-the-art MLLMs, categorized into commercialization models (e.g., Gemini, GPT, Doubao, Kimi, MiMo) and open-source models (e.g., Qwen, InternVL, LLaVA-Video). For video processing, we adopt two distinct input settings: \textit{(1) Visual Frames only (wo sub)}, where models rely solely on visual cues; and \textit{(2) Visual Frames + Subtitles/Audio (w. sub)}, where aligned subtitles, ASR transcripts, or raw audio are provided to supplement visual information. Regarding temporal sampling, we adhere to the specific limitations of each model; for most open-source models, we sample 64 or 512 frames depending on their long-context capabilities, while for commercialization models supporting native video input, we utilize their recommended sampling rates (e.g., 1fps). For Omni-models capable of native audio processing (e.g., MiMo-v2-Omni, Gemini-3-Pro), we feed the raw audio stream in the \textit{w. sub} setting to evaluate their cross-modal listening capabilities, and evaluate them on visual frames only in the \textit{wo sub} setting. All models are evaluated using their recommended settings for the best performance.

\subsection{Benchmark Results}
\begin{table*}[htbp]
\centering
\caption{\textbf{Main Results on the Leaderboard.} 
\textit{w. sub} = with subtitle/audio; \textit{wo sub} = without subtitle/audio (visual frames only). For Omni models that accept raw audio (e.g., MiMo-v2-Omni, Gemini-3-Pro), as well as the human baseline, we report results with raw audio input under the \textit{w. sub} setting, and results with visual frames only under the \textit{wo sub} setting. 
\textit{Sorting}: models are ranked by their \textit{w. sub} Non-Lin Score.
\textit{Frames} = video sampling rate (e.g., 1fps) or maximum sampled frames (e.g., 64, 512). 
\textit{Consistency} = Capability consistency score. 
\textit{Coherence} = Reasoning coherence score. 
\textit{Avg Acc} = average accuracy across all QA pairs.
We argue that the proposed group-based nonlinear scoring (\textbf{Non-Lin Score}) more faithfully reflects true model performance than the traditional per-question average accuracy (Avg Acc).
}
\resizebox{\textwidth}{!}{%
\renewcommand{\arraystretch}{1.3} 
\setlength{\tabcolsep}{3.5pt}      

\begin{tabular}{l c >{\columncolor{subcol}}c c >{\columncolor{subcol}}c c >{\columncolor{subcol}}c c >{\columncolor{subcol}}c c >{\columncolor{subcol}}c c >{\columncolor{subcol}}c c >{\columncolor{subcol}}c c}
\toprule
\multirow{2.5}{*}{\textbf{Model}} & \multirow{2.5}{*}{\textbf{Frames}} & \multicolumn{2}{c}{\textbf{Non-Lin Score}} & \multicolumn{2}{c}{\textbf{Level 1}} & \multicolumn{2}{c}{\textbf{Level 2}} & \multicolumn{2}{c}{\textbf{Level 3}} & \multicolumn{2}{c}{\textbf{Consistency}} & \multicolumn{2}{c}{\textbf{Coherence}} & \multicolumn{2}{c}{\textbf{Avg Acc}} \\
\cmidrule(lr){3-4} \cmidrule(lr){5-6} \cmidrule(lr){7-8} \cmidrule(lr){9-10} \cmidrule(lr){11-12} \cmidrule(lr){13-14} \cmidrule(lr){15-16}
 & & \textbf{w. sub} & wo sub & \textbf{w. sub} & wo sub & \textbf{w. sub} & wo sub & \textbf{w. sub} & wo sub & \textbf{w. sub} & wo sub & \textbf{w. sub} & wo sub & \textbf{w. sub} & wo sub \\
\midrule
\multicolumn{16}{c}{\cellcolor{catgray}\textbf{Human Baseline}} \\
\midrule
Human Expert & - & \textbf{90.7} & - & \textbf{94.8} & - & \textbf{91.1} & - & \textbf{87.9} & - & \textbf{91.7} & - & \textbf{88.9} & - & \textbf{94.9} & - \\\midrule
\multicolumn{16}{c}{\cellcolor{catgray}\textbf{Commercialization Models}} \\
\midrule
Gemini-3-Pro~\cite{gemini3pro} & 1fps & \textbf{49.4} & \textbf{38.2} & \textbf{64.0} & \textbf{43.2} & \textbf{50.0} & \textbf{45.4} & \textbf{40.6} & \textbf{30.2} & \textbf{50.8} & \textbf{39.7} & \textbf{47.0} & \textbf{35.4} & \textbf{66.1} & \textbf{56.8} \\
Doubao-Seed-2.0-Pro-260215~\cite{bytedance2026seed2modelcard} & 1fps & 43.3 & 35.2 & 54.4 & 41.3 & 47.0 & 42.6 & 34.1 & 26.2 & 43.7 & 37.1 & 42.5 & 31.6 & 60.5 & 53.1 \\
Gemini-3-Flash~\cite{gemini3pro} & 1fps & 42.5 & 32.9 & 58.3 & 41.5 & 44.8 & 37.4 & 31.7 & 25.0 & 44.9 & 35.6 & 38.2 & 28.0 & 61.1 & 52.4 \\
MiMo-v2-Omni~\cite{coreteam2025mimovltechnicalreport} & 1fps & 38.6 & 29.9 & 52.6 & 38.7 & 43.1 & 36.0 & 27.4 & 20.4 & 39.7 & 32.4 & 36.6 & 25.4 & 56.1 & 47.1 \\
GPT-5~\cite{gpt5} & 50 & 37.0 & 26.4 & 44.5 & 32.2 & 39.1 & 28.6 & 31.1 & 21.3 & 37.4 & 28.1 & 36.3 & 23.2 & 55.6 & 44.7 \\
Kimi-K2.5~\cite{team2025kimi} & 64 & 36.1 & 27.3 & 44.3 & 30.5 & 40.0 & 32.8 & 28.5 & 21.6 & 36.0 & 28.7 & 36.2 & 24.9 & 54.4 & 46.0 \\

\midrule
\multicolumn{16}{c}{\cellcolor{catgray}\textbf{Open-Source Models (Instruct)}} \\
\midrule
Qwen3-VL-235B-A22B-Instruct~\cite{bai2025qwen3} & 64 & \textbf{25.0} & 16.5 & \textbf{30.7} & \textbf{20.6} & 25.2 & 16.8 & 21.6 & 13.9 & \textbf{25.0} & 17.7 & 25.0 & 14.2 & \textbf{43.3} & 33.8 \\
Qwen3.5-397B-A17B-Instruct~\cite{qwen3.5} & 64 & 24.5 & \textbf{16.9} & 27.3 & 16.8 & \textbf{25.9} & \textbf{18.1} & \textbf{21.9} & \textbf{16.1} & 23.9 & \textbf{18.2} & \textbf{25.7} & \textbf{14.6} & 42.1 & \textbf{35.0} \\
Qwen3.5-27B-Instruct~\cite{qwen3.5} & 64 & 23.9 & 14.0 & 28.9 & 15.8 & 23.2 & 14.8 & 21.5 & 12.3 & 23.4 & 14.8 & 24.8 & 12.6 & 41.9 & 31.7 \\
InternVL3\_5-241B-A28B-Instruct~\cite{wang2025internvl3} & 64 & 23.1 & 15.8 & 28.2 & 18.1 & 23.7 & 16.3 & 19.6 & 14.1 & 23.4 & 17.2 & 22.5 & 13.2 & 41.4 & 32.9 \\
Qwen2-5-VL-72B-Instruct~\cite{bai2025qwen25vltechnicalreport} & 64 & 22.1 & 13.7 & 25.8 & 15.3 & 21.2 & 13.7 & 20.6 & 12.6 & 20.9 & 14.1 & 24.3 & 12.9 & 38.9 & 30.3 \\
Qwen3.5-122B-A10B-Instruct~\cite{qwen3.5} & 64 & 20.9 & 12.5 & 24.9 & 14.8 & 21.8 & 12.9 & 17.9 & 10.9 & 20.1 & 13.3 & 22.4 & 11.1 & 39.2 & 29.5 \\
Qwen3.5-35B-A3B-Instruct~\cite{qwen3.5} & 64 & 19.9 & 11.3 & 24.2 & 13.7 & 18.6 & 11.3 & 18.4 & 9.9 & 18.4 & 12.4 & 22.7 & 9.4 & 36.9 & 28.6 \\
KimiVL-16B-A3B-Instruct~\cite{team2025kimi} & 64 & 19.0 & 15.0 & 23.9 & 18.7 & 17.6 & 13.5 & 17.1 & 13.8 & 19.4 & 16.7 & 18.2 & 11.7 & 37.3 & 32.3 \\
Qwen3.5-9B-Instruct~\cite{qwen3.5} & 64 & 18.3 & 10.3 & 20.7 & 11.5 & 17.6 & 9.9 & 17.4 & 10.0 & 16.7 & 10.6 & 21.2 & 9.9 & 35.5 & 26.0 \\
Qwen3-VL-8B-Instruct~\cite{bai2025qwen3} & 64 & 18.2 & 12.4 & 21.8 & 15.4 & 17.0 & 12.5 & 16.9 & 10.7 & 17.1 & 13.2 & 20.3 & 11.1 & 35.7 & 27.9 \\
Qwen3-VL-4B-Instruct~\cite{bai2025qwen3} & 64 & 17.6 & 11.6 & 23.6 & 16.4 & 14.6 & 9.5 & 16.2 & 10.3 & 16.3 & 11.5 & 19.9 & 11.7 & 34.1 & 26.2 \\
LLaVA-Video-72B-Qwen2~\cite{zhang2024llava} & 64 & 17.2 & 11.3 & 21.8 & 14.8 & 14.8 & 11.1 & 16.3 & 9.5 & 17.1 & 12.3 & 17.5 & 9.5 & 34.4 & 27.3 \\
Qwen3-Omni-30B-A3B-Instruct~\cite{xu2025qwen3} & 64 & 17.1 & 10.9 & 21.7 & 15.0 & 15.0 & 10.2 & 16.0 & 8.9 & 16.5 & 11.8 & 18.3 & 9.2 & 34.1 & 27.0 \\
Qwen3-VL-30B-A3B-Instruct~\cite{bai2025qwen3} & 64 & 16.8 & 7.8 & 18.0 & 9.3 & 17.6 & 7.7 & 15.6 & 7.1 & 15.7 & 8.5 & 19.0 & 6.7 & 33.2 & 21.3 \\
InternVL3\_5-30B-A3B-Instruct~\cite{wang2025internvl3} & 64 & 16.6 & 12.6 & 21.6 & 17.5 & 14.5 & 11.0 & 15.2 & 10.9 & 15.9 & 13.7 & 18.0 & 10.6 & 34.2 & 28.7 \\
InternVL3\_5-8B-Instruct~\cite{wang2025internvl3} & 64 & 16.4 & 11.4 & 21.6 & 14.8 & 15.3 & 10.8 & 14.2 & 9.8 & 16.1 & 12.4 & 17.1 & 9.4 & 32.4 & 26.0 \\
Qwen3.5-4B-Instruct~\cite{qwen3.5} & 64 & 15.4 & 9.5 & 17.6 & 11.9 & 16.1 & 9.0 & 13.5 & 8.5 & 14.6 & 9.8 & 16.7 & 9.1 & 32.3 & 24.1 \\
MiMo-VL-7B-SFT-2508~\cite{coreteam2025mimovltechnicalreport} & 64 & 14.7 & 10.7 & 19.8 & 14.3 & 13.2 & 9.6 & 12.8 & 9.3 & 14.1 & 11.2 & 15.8 & 9.7 & 31.2 & 24.6 \\
Qwen2-5-VL-7B-Instruct~\cite{bai2025qwen25vltechnicalreport} & 64 & 14.4 & 10.3 & 20.0 & 14.7 & 13.0 & 9.3 & 12.1 & 8.4 & 13.7 & 11.1 & 15.6 & 8.8 & 30.9 & 25.3 \\
VideoLLaMA3-7B~\cite{zhang2025videollama} & 64 & 14.2 & 10.9 & 19.0 & 15.8 & 12.7 & 9.2 & 12.4 & 9.1 & 14.3 & 12.3 & 13.9 & 8.2 & 30.6 & 25.7 \\
InternVL-3-5-4B-Instruct~\cite{wang2025internvl3} & 64 & 14.2 & 9.9 & 21.6 & 13.5 & 10.5 & 9.2 & 12.6 & 8.2 & 14.1 & 10.7 & 14.6 & 8.4 & 30.5 & 24.6 \\
Keye-VL-1-5-8B~\cite{yang2025kwai} & 64 & 14.1 & 8.9 & 19.1 & 12.2 & 12.4 & 8.4 & 12.4 & 7.3 & 13.7 & 10.4 & 14.8 & 6.0 & 31.1 & 23.8 \\
VITA-1.5-7B~\cite{fu2025vita} & 16 & 11.3 & 8.2 & 14.1 & 11.6 & 9.4 & 7.9 & 11.0 & 6.5 & 10.3 & 8.9 & 13.0 & 7.1 & 25.8 & 21.8 \\
LLaVA-Video-7B-Qwen2~\cite{zhang2024llava} & 64 & 9.7 & 7.2 & 15.9 & 12.7 & 7.4 & 5.6 & 7.5 & 5.1 & 9.9 & 8.5 & 9.0 & 4.8 & 24.0 & 19.9 \\

\midrule
\multicolumn{16}{c}{\cellcolor{catgray}\textbf{Open-Source Models (Thinking)}} \\
\midrule
Qwen3.5-397B-A17B-Think~\cite{qwen3.5} & 512 & \textbf{39.1} & \textbf{30.3} & \textbf{50.3} & \textbf{35.4} & \textbf{41.8} & \textbf{36.5} & \textbf{30.7} & \textbf{22.9} & \textbf{39.0} & \textbf{32.2} & \textbf{39.4} & \textbf{26.9} & \textbf{55.9} & \textbf{48.8} \\
Qwen3.5-27B-Think~\cite{qwen3.5} & 512 & 31.4 & 19.5 & 34.4 & 22.3 & 36.4 & 23.1 & 26.2 & 14.4 & 31.5 & 20.6 & 31.3 & 16.5 & 49.6 & 37.4 \\
Qwen3.5-397B-A17B-Think~\cite{qwen3.5} & 64 & 30.6 & 20.7 & 34.8 & 20.7 & 30.4 & 23.1 & 28.4 & 19.0 & 29.5 & 20.9 & 32.7 & 20.4 & 48.9 & 39.3 \\
Qwen3.5-27B-Think~\cite{qwen3.5} & 64 & 29.6 & 17.6 & 35.0 & 20.0 & 30.6 & 18.5 & 25.8 & 15.5 & 28.5 & 18.2 & 31.7 & 16.4 & 47.6 & 34.5 \\
Qwen3-VL-235B-A22B-Think~\cite{bai2025qwen3} & 512 & 28.1 & 19.0 & 32.6 & 21.4 & 30.3 & 23.4 & 23.9 & 14.5 & 28.3 & 20.4 & 27.8 & 16.5 & 47.2 & 36.8 \\
Qwen3.5-122B-A10B-Think~\cite{qwen3.5} & 64 & 26.7 & 16.3 & 30.1 & 19.0 & 27.9 & 16.0 & 23.9 & 14.9 & 26.4 & 16.8 & 27.3 & 15.3 & 45.1 & 33.0 \\
Qwen3.5-Omni-Plus~\cite{qwen35omniblog} & 64 & 25.6 & 17.3 & 31.2 & 19.6 & 26.6 & 18.2 & 21.6 & 15.2 & 25.3 & 17.3 & 26.0 & 17.2 & 44.3 & 35.0 \\
Qwen3-VL-235B-A22B-Think~\cite{bai2025qwen3} & 64 & 26.3 & 15.9 & 32.0 & 19.1 & 27.5 & 16.8 & 22.3 & 13.4 & 26.0 & 17.2 & 27.0 & 13.4 & 44.9 & 33.3 \\
Qwen3.5-9B-Think~\cite{qwen3.5} & 512 & 26.2 & 19.5 & 30.5 & 21.8 & 30.1 & 24.5 & 20.9 & 14.6 & 25.8 & 21.0 & 26.8 & 16.8 & 44.5 & 36.8 \\
Qwen3.5-35B-A3B-Think~\cite{qwen3.5} & 64 & 25.8 & 15.6 & 31.3 & 18.2 & 24.9 & 15.8 & 23.2 & 13.9 & 24.9 & 16.1 & 27.5 & 14.6 & 43.9 & 32.6 \\
Qwen3.5-122B-A10B-Think~\cite{qwen3.5} & 512 & 25.1 & 14.0 & 27.4 & 15.8 & 25.6 & 14.0 & 23.4 & 12.9 & 24.3 & 14.5 & 26.5 & 12.9 & 43.4 & 30.6 \\
Qwen3.5-35B-A3B-Think~\cite{qwen3.5} & 512 & 23.9 & 12.2 & 29.5 & 13.9 & 24.5 & 11.4 & 20.1 & 11.7 & 23.4 & 12.8 & 24.7 & 11.1 & 41.5 & 28.5 \\
Qwen3.5-9B-Think~\cite{qwen3.5} & 64 & 23.2 & 13.7 & 28.3 & 16.7 & 24.7 & 12.6 & 19.0 & 12.6 & 23.6 & 13.9 & 22.2 & 13.1 & 40.3 & 28.9 \\
Qwen3-VL-30B-A3B-Think~\cite{bai2025qwen3} & 512 & 22.1 & 13.2 & 27.6 & 16.9 & 22.4 & 14.7 & 18.6 & 10.0 & 21.7 & 14.8 & 22.8 & 10.4 & 39.8 & 29.5 \\
GLM4.5V-Think~\cite{hong2025glm} & 64 & 21.8 & 13.3 & 26.2 & 14.9 & 21.7 & 14.7 & 19.4 & 11.4 & 20.5 & 14.7 & 24.3 & 10.6 & 39.2 & 29.6 \\Qwen3-VL-30B-A3B-Think~\cite{bai2025qwen3} & 64 & 21.1 & 13.5 & 27.0 & 17.4 & 21.1 & 13.7 & 17.6 & 11.1 & 20.2 & 14.1 & 22.6 & 12.5 & 38.6 & 29.4 \\
Qwen3.5-4B-Think~\cite{qwen3.5} & 64 & 20.7 & 11.6 & 26.2 & 14.9 & 21.0 & 11.1 & 17.3 & 9.9 & 20.4 & 12.4 & 21.4 & 10.0 & 38.8 & 26.5 \\
Qwen3-VL-8B-Think~\cite{bai2025qwen3} & 512 & 20.2 & 14.4 & 23.5 & 17.0 & 21.4 & 16.7 & 17.4 & 11.1 & 20.1 & 15.1 & 20.3 & 13.0 & 37.3 & 29.7 \\
GLM4.6-Flash~\cite{hong2025glm} & 64 & 19.8 & 12.7 & 26.8 & 17.3 & 19.4 & 13.4 & 15.9 & 9.5 & 19.6 & 14.0 & 20.0 & 10.3 & 37.4 & 27.7 \\
Qwen3-VL-8B-Think~\cite{bai2025qwen3} & 64 & 19.7 & 11.8 & 23.9 & 13.1 & 19.2 & 11.8 & 17.5 & 11.0 & 19.2 & 12.7 & 20.5 & 10.0 & 37.2 & 27.0 \\
Qwen3-Omni-30B-A3B-Think~\cite{xu2025qwen3} & 64 & 19.5 & 12.6 & 24.7 & 15.8 & 17.5 & 12.3 & 17.9 & 11.0 & 18.6 & 13.3 & 21.2 & 11.4 & 36.5 & 28.6 \\
MiMo-VL-7B-RL-2508~\cite{coreteam2025mimovltechnicalreport} & 64 & 19.4 & 11.9 & 24.9 & 16.1 & 18.8 & 12.5 & 16.6 & 9.0 & 18.9 & 12.5 & 20.3 & 10.8 & 36.0 & 27.0 \\
Qwen3-VL-4B-Think~\cite{bai2025qwen3} & 64 & 18.2 & 11.6 & 23.3 & 15.0 & 16.6 & 11.3 & 16.3 & 9.8 & 17.9 & 12.6 & 18.6 & 9.7 & 35.3 & 26.2 \\
KimiVL-16B-A3B-Think~\cite{team2025kimi} & 64 & 15.7 & 11.7 & 21.8 & 16.4 & 14.0 & 10.6 & 13.3 & 9.7 & 14.9 & 12.7 & 17.1 & 9.9 & 32.4 & 27.3 \\
GLM4.1V-Think~\cite{hong2025glm} & 64 & 14.0 & 9.4 & 18.8 & 14.1 & 13.2 & 7.5 & 11.8 & 8.1 & 13.8 & 10.6 & 14.4 & 7.3 & 30.3 & 23.7 \\
\bottomrule
\end{tabular}%
}
\label{tab:main_leaderboard}
\end{table*}

Based on the results presented in Table~\ref{tab:main_leaderboard}, we offer the following interpretations regarding the current state of video understanding:

\textbf{Significant Gap with Human Performance.}
We establish a rigorous human expert baseline, where human annotators achieve an outstanding Non-Lin Score of \textbf{90.7} and an average accuracy of \textbf{94.9\%}. In stark contrast, the best-performing commercialization model, Gemini-3-Pro, only reaches a Non-Lin Score of 49.4. This massive 41.3-point performance gap in the group-based nonlinear metric demonstrates that while current MLLMs show promise on isolated queries, they are still far from achieving true human-level reliability and sustained logical coherence in complex, long-horizon video reasoning.

\textbf{Hierarchical Bottlenecks in Video Understanding.}
Under our progressive evaluation framework, Level~1 (multi-point information aggregation) serves as the foundation for Level~2 (temporal understanding), which in turn underpins Level~3 (complex reasoning). Consistent with this hierarchical dependency, we observe a monotonic performance degradation from Level~1 to Level~3 across all models.

Importantly, the relatively low performance at Level~3 is not solely attributable to insufficient high-level reasoning capability. It is also fundamentally constrained by weaknesses in the lower-level competencies. Errors in Level~1 propagate to Level~2, leading to flawed temporal modeling, which ultimately undermines the reliability of multi-hop reasoning and global consistency required at Level~3.

This cascading effect highlights that improving complex reasoning is not merely a matter of scaling reasoning modules, but requires a holistic enhancement of the entire capability stack, particularly strengthening perception and temporal grounding as prerequisite components.

\textbf{Commercialization Dominance.} 
Gemini-3-Pro achieves a Non-Lin Score of 49.4, significantly outperforming the best open-source model, Qwen3.5-397B Think (39.1). A notable observation is the performance under the \textit{wo sub} setting: commercialization models such as Doubao-Seed-2.0-Pro and GPT-5 remain competitive with, or even exceed, most open-source models under the \textit{w. sub} setting (e.g., surpassing the best open-source Instruct model, Qwen3-VL-235B-A22B-Instruct at 25.0). In contrast, most open-source models show much larger drops once subtitle/audio side information is removed. This pattern suggests that current open-source video MLLMs are generally more dependent on textual or auxiliary multimodal cues, while commercialization models appear more robust to their removal.

\textbf{The Benefit of Native Audio in Omni Architectures.}
For Omni models capable of processing raw audio (e.g., MiMo-v2-Omni, Gemini-3-Pro), the \textit{w. sub} setting reflects native audio-visual fusion rather than text-only subtitle input. Under this setting, direct audio integration brings clear gains: Gemini-3-Pro improves from 38.2 to 49.4 (+11.2), and MiMo-v2-Omni improves from 29.9 to 38.6 (+8.7). These results suggest that native audio can provide complementary semantic and paralinguistic information beyond visual frames alone, and can help reduce reliance on purely textual subtitle/ASR pipelines.

\textbf{High Performance of Small-Scale Models.}
Model scale is not the sole determinant of performance. Notably, Qwen3.5-27B-Think achieves a Non-Lin Score of 31.4, outperforming significantly larger models such as Qwen3-VL-235B-Instruct (25.0) and matching or exceeding many 72B-class baselines. Similarly, Qwen3.5-9B-Think reaches 26.2, showing that a relatively small model can remain competitive when paired with strong post-training, long-context support, and reasoning alignment.
These results highlight that model quality—particularly training recipe, data curation, and alignment techniques—can have a significant impact on performance, in some cases outweighing the advantages of sheer parameter scale.

\subsection{Analysis Experiments}

\subsubsection{Advantage of Group-Based Nonlinear Scoring (Non-Lin Score)}
\label{subsubsec:nonlin_vs_avgacc}

We compare two evaluation metrics: our proposed \textbf{group-based nonlinear score} (\textbf{Non-Lin Score}) and the \textbf{per-question average accuracy} (\textbf{Avg Acc}).

\textbf{Intra-Model Comparison.}
Gemini-3-Pro and Doubao-Seed-2.0-Pro achieve Avg Acc of 66.1\% and 60.5\%, respectively, which may appear relatively high under conventional per-question accuracy. However, under the group-based nonlinear scoring, their Non-Lin Scores drop to 49.4 and 43.3. This discrepancy indicates that even state-of-the-art models often fail to consistently answer all correlated questions within the same group.
By explicitly leveraging the group structure, our nonlinear scoring is less sensitive to isolated correct predictions and instead emphasizes consistency across related queries, thereby providing a more faithful assessment of true model capability.

\textbf{Inter-Model Comparison via Robustness Ratio.}
The ratio \textbf{Non-Lin Score / Avg Acc} quantifies the performance degradation from “single-question correctness” to “within-group consistency,” serving as an indicator of model robustness. Gemini-3-Pro and Doubao-Seed-2.0-Pro achieve ratios of approximately 75\% and 72\%, respectively, while Qwen3.5-397B (512 frames) attains around 70\%. In contrast, smaller models such as LLaVA-Video-7B exhibit substantially lower ratios, around 40
A lower ratio indicates that a model more frequently answers only a subset of questions correctly within a group, suggesting weaker consistency and reduced robustness across correlated queries.

\begin{table}[h]
    \centering
    \small
    \begin{tabular}{l c c c}
        \toprule
        \textbf{Model} & \textbf{Avg Acc (\%)} & \textbf{Non-Lin Score (Ours)} & \textbf{Non-Lin / Avg Acc} \\
        \midrule
        Gemini-3-Pro & 66.1 & 49.4 & 74.7\% \\
        Doubao-Seed-2.0-Pro-260215 & 60.5 & 43.3 & 71.6\% \\
        Gemini-3-Flash & 61.1 & 42.5 & 69.6\% \\
        Qwen3.5-397B-A17B-Think (512) & 55.9 & 39.1 & 69.9\% \\
        MiMo-v2-Omni & 56.1 & 38.6 & 68.8\% \\
        GPT-5 & 55.6 & 37.0 & 66.5\% \\
        Qwen3.5-27B-Think (512) & 49.6 & 31.4 & 63.3\% \\
        LLaVA-Video-7B-Qwen2 & 24.0 & 9.7 & 40.4\% \\
        \bottomrule
    \end{tabular}
    \caption{\textbf{Avg Acc vs.\ Non-Lin Score.} The gap and the ratio reflect robustness: whether a model can answer multiple correlated questions correctly within the same group.}
    \label{tab:avgacc_vs_nonlin}
\end{table}

\subsubsection{Analysis of Capability Consistency and Reasoning Coherence}
\label{subsubsec:consistency_coherence}

\begin{figure}[h]
    \centering
    \includegraphics[width=\linewidth]{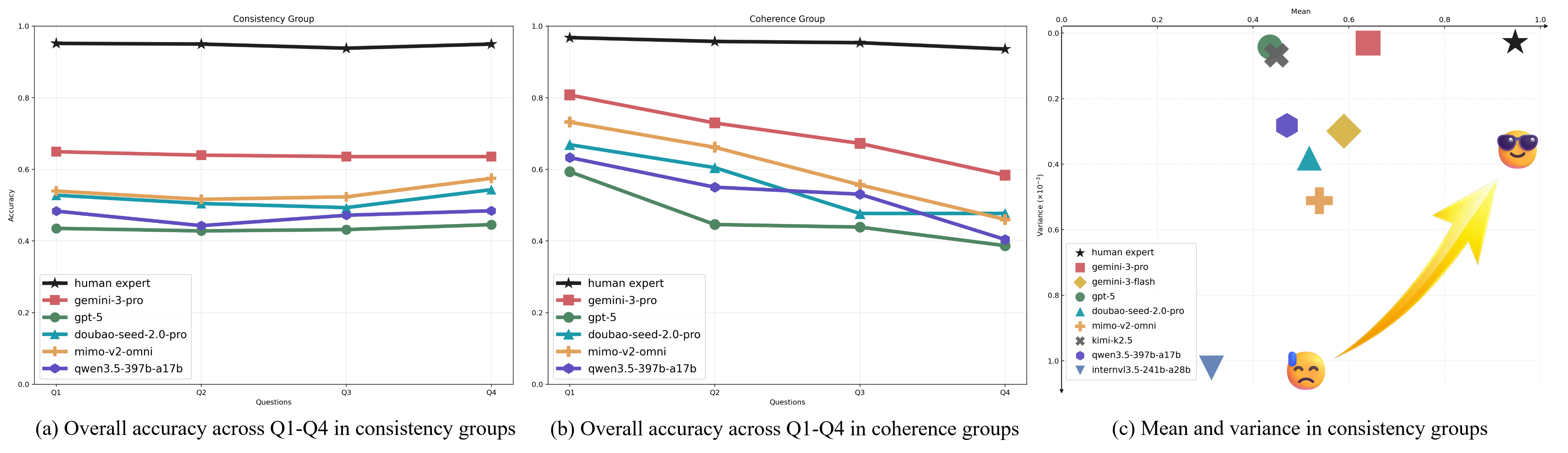}
    \caption{\textbf{Q1--Q4 Accuracy Trends and Stability.} Trends under (a) Capability consistency, (b) Reasoning coherence, and (c) mean/variance statistics under capability consistency.}
    \label{fig:q1_q4_accuracy}
\end{figure}

We analyze Q1--Q4 overall accuracy trends of five representative models, including Gemini-3-Pro, GPT-5, Doubao-Seed-2.0-Pro, MiMo-v2-Omni, and Qwen3.5-397B-A17B, under two group types: \textit{capability consistency} in \Cref{fig:q1_q4_accuracy} (a) and \textit{reasoning coherence} in \Cref{fig:q1_q4_accuracy} (b).

\textbf{Data Perspective.}
In \textit{capability consistency} groups, model accuracies across Q1--Q4 remain broadly comparable, suggesting that the difficulty is relatively balanced across question indices. In contrast, within \textit{reasoning coherence} groups, all models exhibit a clear monotonic decline from Q1 to Q4, reflecting an intentionally designed difficulty progression with increasingly strict logical dependencies along the reasoning chain.

\textbf{Model Perspective.}
In \textit{capability consistency} groups, Gemini-3-Pro and GPT-5 show the smallest fluctuation across Q1--Q4, indicating strong stability.
In \textit{reasoning coherence} groups, stronger models exhibit a smooth decline in accuracy from Q1 to Q4 as question difficulty increases, whereas weaker models show more irregular patterns.
One possible explanation is that stronger models are more sensitive to incremental changes in question difficulty, resulting in a more uniform degradation as reasoning depth increases. In contrast, weaker models tend to exhibit higher stochasticity, leading to unstable performance across progressively harder questions.

\textbf{Mean--Variance Analysis on Capability Consistency.}
In \Cref{fig:q1_q4_accuracy} (c), we further summarize the mean and variance of overall Q1--Q4 accuracy for eight models in the \textit{capability consistency} setting. Mean reflects average capability while variance reflects stability/robustness. Two findings emerge: (1) Gemini-3-Pro achieves both the highest mean accuracy and the lowest variance, indicating the strongest overall performance and stability, while GPT-5 and Kimi-K2.5 follow closely behind in terms of stability; (2) commercialization models generally outperform open-source models, yet all models still remain substantially below human performance, indicating a significant gap to close.

\subsubsection{Effect of Thinking Mode on Video-MME-v2}
\label{subsubsec:thinking_effect}

\begin{figure}[h]
    \centering
    \includegraphics[width=\linewidth]{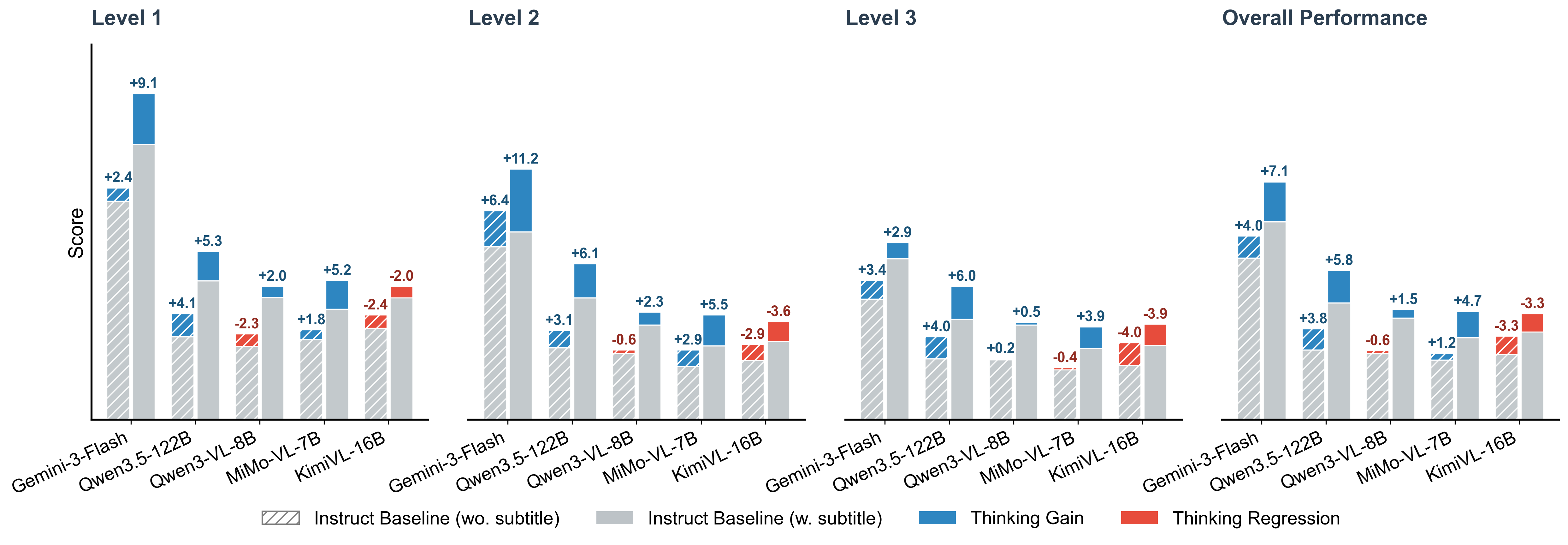}
    \caption{\textbf{Effect of Thinking Mode on Video-MME-v2.} Performance changes induced by enabling \emph{Thinking} for instruction-tuned baseline models, evaluated under both \emph{wo. subtitle} and \emph{w. subtitle} settings.}
    \label{fig:thinking_effect_v2}
\end{figure}

In \Cref{fig:thinking_effect_v2}, we compare the performance changes of instruction-tuned baseline models after enabling the \emph{Thinking} mode, under both \emph{wo. subtitle} and \emph{w. subtitle} settings. Gemini-3-Flash is a special case due to model constraints. For this model, we instead compare its \textit{minimal thinking} mode with the standard thinking mode.

\textbf{Text Modality Helps Unlock Reasoning.}
Overall, enabling \emph{Thinking} leads to larger performance gains in the \emph{w. subtitle} setting than in \emph{wo. subtitle}, suggesting that explicit textual semantics serve as strong anchors for multi-step reasoning. For instance, Qwen3.5-122B-A10B-Think (64 frames) achieves improvements of \textbf{+3.8/+5.8} (\emph{wo./w. subtitle}), demonstrating that subtitles can substantially amplify the effectiveness of reasoning mechanisms.

\textbf{Thinking Can Also Cause Regression.}
We also observe noticeable performance regressions after enabling \emph{Thinking} for certain models, often more pronounced in the \emph{wo. subtitle} setting. For instance, Qwen3-VL-8B shows a drop of \textbf{-0.6} (\emph{wo. subtitle}), while KimiVL-16B degrades by \textbf{-3.3/-3.3} overall. Notably, at the most reasoning-intensive Level~3, KimiVL-16B further declines by \textbf{-4.0/-3.9}.
These observations suggest that current reasoning mechanisms remain imperfect and may introduce additional noise, particularly in settings where textual modality is absent.

\subsubsection{Overall Model Capability Analysis on Video-MME-v2}
\label{subsubsec:capability_analysis_v2}

Following the three task levels of Video-MME-v2, we abstract three core capabilities required for strong performance.
\textbf{C1}: Omni-modal information aggregation;
\textbf{C2}: long-range temporal/long-context understanding;
\textbf{C3}: complex reasoning.
We tag representative models with these capabilities and compare their Non-Lin Scores in Table~\ref{tab:capability_analysis_videommev2}

\textbf{Synergy of Multi-Dimensional Core Capabilities.}
Scores tend to correlate with how complete the capability profile is: models with C1+C2+C3 together generally perform better. For example, Gemini-3-Pro has a relatively complete profile and scores 49.4; Gemini-3-Flash follows with 42.5. MiMo-v2-Omni, also tagged with C1+C2+C3, reaches 38.6. This suggests that in complex video understanding, the synergy of omni-modal perception, long-horizon temporal modeling, and deep reasoning may be an important factor for overall performance.

\textbf{Model Scale and Capability Compensation.}
Besides capability combination, results show that scale has a significant effect on base performance: larger parameter count can partly compensate for missing capabilities. For example, Qwen3.5-397B-A17B-Think (512 frames), tagged mainly with C2 and C3, reaches 39.1 and slightly surpasses MiMo-v2-Omni (38.6), which is tagged with C1+C2+C3. This suggests that when scale increases substantially, the model’s overall capacity can partly offset an incomplete capability profile, even though more complete capability coverage remains advantageous overall.

\textbf{Impact of Frame Count on Performance.}
For the same model, increasing frame count can significantly improve performance. For example, Qwen3.5-397B-A17B-Think with 512 frames scores 39.1, while with 64 frames it scores only 30.6—an 8.5-point gap. This highlights the importance of long-context processing capability (C2) for complex video understanding tasks.

\begin{table}[h]
    \centering
    \small
    \setlength{\tabcolsep}{6pt}
    \begin{tabular}{l c c c c}
        \toprule
        \textbf{Model} & \textbf{Non-Lin Score} & \textbf{C1} & \textbf{C2} & \textbf{C3} \\
        \midrule
        Gemini-3-Pro                     & 49.4 & \cmark & \cmark & \cmark \\
        Gemini-3-Flash                   & 42.5 & \cmark & \cmark & \cmark \\
        Qwen3.5-397B-A17B-Think (512)    & 39.1 &        & \cmark & \cmark \\
        MiMo-v2-Omni                     & 38.6 & \cmark & \cmark & \cmark \\
        Qwen3.5-397B-A17B-Think (64)     & 30.6 &        & \cmark & \cmark \\
        Qwen3-VL-235B-A22B-Think (512)   & 28.1 &        & \cmark & \cmark \\
        Qwen3-Omni-30B-A3B-Think         & 19.5 & \cmark & \cmark & \cmark \\
        Qwen3-Omni-30B-A3B-Instruct      & 17.1 & \cmark & \cmark &        \\
        \bottomrule
    \end{tabular}
    \caption{\textbf{Capability Profiling on Video-MME-v2.} C1: omni-modal joint perception; C2: long-context/long-range temporal modeling; C3: complex reasoning (Thinking).}
    \label{tab:capability_analysis_videommev2}
\end{table}

\subsubsection{Capability Radar Analysis}
\label{subsubsec:capability_radar}

\begin{figure}[h]
    \centering
    \includegraphics[width=\linewidth]{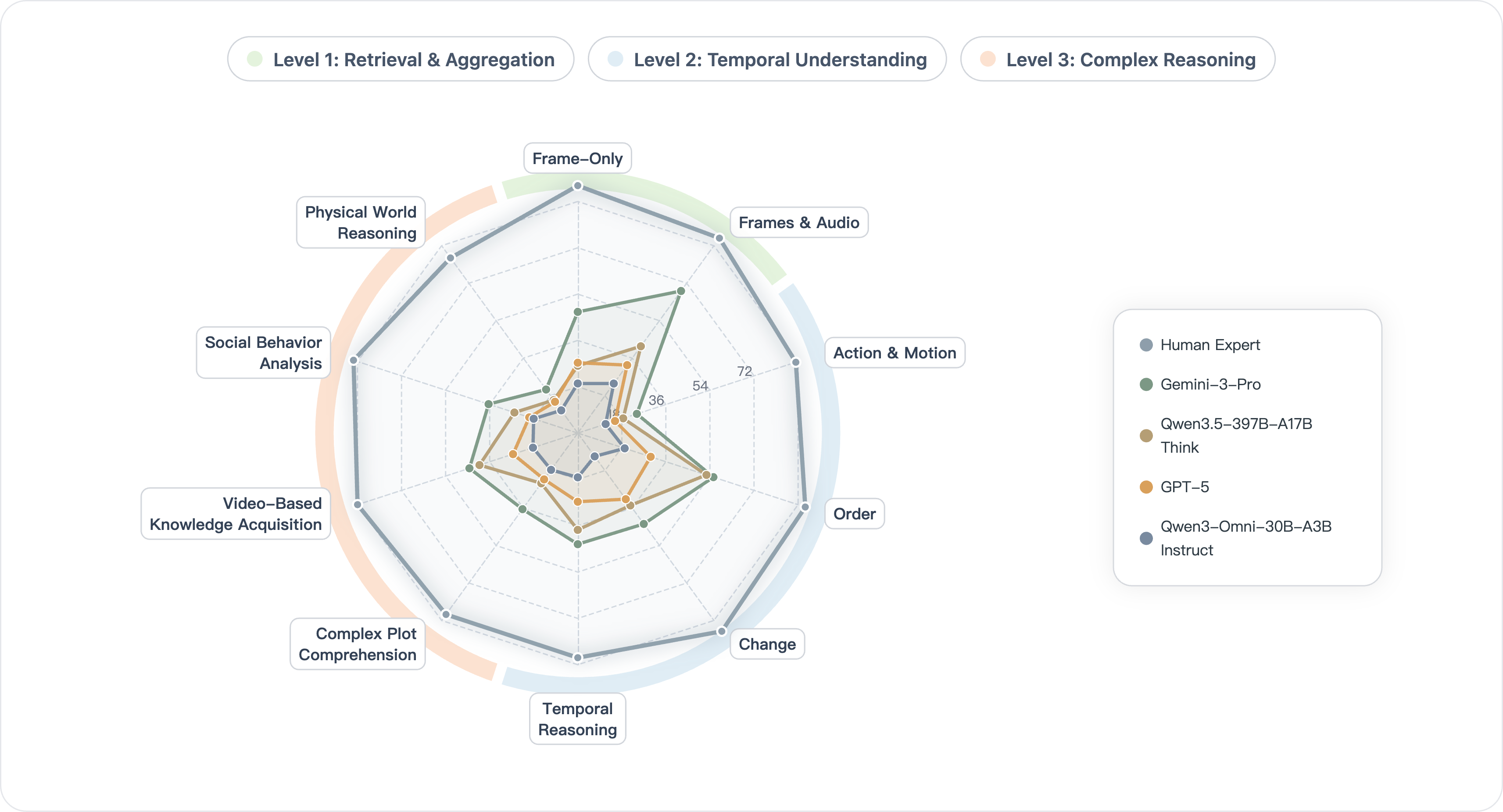}
    \caption{\textbf{Capability Radar Across Video-MME-v2 Dimensions.} For a wider range of models, please visit our project page, where you can select to view the radar chart performance of different models.}    
    \label{fig:capability_radar}
\end{figure}

In \Cref{fig:capability_radar}, we compare the models based on the capability dimensions defined by Video-MME-v2. Three major observations are summarized.

\textbf{Substantial Gains from Audio.}
Gemini-3-Pro shows a clear peak on the \emph{Frames\&Audio} dimension, suggesting stronger cross-modal alignment and integration when processing synchronized vision+audio, compared to models that rely more heavily on visual frames.

\textbf{Long-Horizon Temporal Reasoning Advantage.}
On dimensions requiring long-video temporal modeling and cross-segment inference (e.g., \emph{Order} and \emph{Video-Based Knowledge Acquisition}), Gemini-3-Pro also maintains a noticeable lead, indicating more robust long-context processing and temporal modeling.

\textbf{Significant Headroom Remains.}
Even for SOTA models, scores on challenging dimensions such as \emph{Action \& Motion} and \emph{Physical World Reasoning} remain below 30, highlighting persistent limitations in fine-grained action semantics and physical-law reasoning.

\section{Conclusion}
In this paper, we introduce Video-MME-v2, a benchmark designed to comprehensively evaluate the robustness and faithfulness of video MLLMs. We propose a progressive multi-level hierarchy that spans diverse video understanding tasks. Complementing this hierarchy, we develop a group-based evaluation framework, where consistency-based groups probe capability robustness and coherence-based groups assess reasoning faithfulness. To further sharpen the evaluation, we introduce a non-linear scoring scheme applied within each group.
Through extensive experiments, we demonstrate the challenging nature of Video-MME-v2, validate the effectiveness of our group-based evaluation and scoring strategy, and provide in-depth analysis of model capabilities and reasoning behaviors. Overall, Video-MME-v2 establishes a new, more rigorous testbed that pushes frontier video MLLMs toward more robust and faithful video understanding.

\section*{Contributors and Acknowledgements}

\vspace{0.5em}
\textbf{Project Lead} \\
Chaoyou Fu (bradyfu24@gmail.com)

\vspace{1em}
\textbf{Contributors} \\
* indicates core contributors with equal contributions. \\[0.5em]
Chaoyou Fu$^\ast$, Haozhi Yuan$^\ast$, Yuhao Dong$^\ast$, Yi-Fan Zhang$^\ast$, Yunhang Shen, Xiaoxing Hu, Xueying Li, Jinsen Su, Chengwu Long, Xiaoyao Xie, Yongkang Xie, Xiawu Zheng, Xue Yang, Haoyu Cao, Yunsheng Wu, Ziwei Liu

\vspace{1em}
\textbf{Senior Leads} \\
Xing Sun, Caifeng Shan, Ran He

\vspace{1em}
\textbf{Acknowledgements} \\
We sincerely thank Yang Shi, Ningjing Liu, Weifan Fang, Wei Zhang, Xiang Liu, Lijiang Li, Ruoliu Yang, Jinqi Wu, Yiming Zhong, Xianyun Sun, Ruohan Liu, Yonghui Niu, Feiyang Duan, Xinyue Cai, Yan Yang, Qingyuan Cao, Zijian Liu, Shaoqi Dong, Xihang Qiu, Dong Chen, Yiyang Hu, Zhenyu Huang, Kaifeng Lin, Haodong Liang, Ziming Shan, Tian Wu, Hao Pan, Ziqian Jiang, Derui Mi, Wangwang Tan, Yanglu Zeng, Mu Wang, Maike Yuan, Yushen Wang, Siyuan Huang, Tianze Jie, Senji Liu, Duxiu Huang, Ziyang Gong, Xiaolin Liu, Haiyao Jin, Junwei Luo, Ruilin Li, Jiale Li, Yiguo He, Ziqian Fan, Feixiang Ruan, Ziqi Ye, Qihao Yang, Jiahao Wang, Zhefeng Qv, Jiayun Li, Keyu Chen, Wei Chen, Zhiqiang Lu, Zirun Zhu for their support of Video-MME-v2 project.

\bibliography{example_paper}
\bibliographystyle{plain}

\newpage
\appendix
\begingroup
\small
\setlength{\tabcolsep}{6pt}
\begin{longtable}{p{0.06\linewidth} p{0.22\linewidth} p{0.28\linewidth} p{0.38\linewidth}}

\caption{Video-MME v2 leaf-level task definitions (full version). This table provides detailed, reproducible descriptions for every leaf category in our Level~1--3 taxonomy, including the intended evidence type (visual/audio), whether temporal order is required, and the type of inference expected. These definitions are used as annotation guidelines so that each Group-QA can be assigned to a single leaf category with clear boundaries, enabling fine-grained analysis of capability consistency and reasoning coherence across models.}
\label{tab:taxonomy_full} \\

\toprule
\textbf{Lv} & \textbf{Parent Category} & \textbf{Leaf Category} & \textbf{Definition (detailed)} \\
\midrule
\endfirsthead

\caption[]{Video-MME v2 leaf-level task definitions (Continued)} \\
\toprule
\textbf{Lv} & \textbf{Parent Category} & \textbf{Leaf Category} & \textbf{Definition (detailed)} \\
\midrule
\endhead

\midrule
\multicolumn{4}{r}{\textit{Continued on next page...}} \\
\endfoot

\bottomrule
\endlastfoot

L1 & Frame Only & Visual Recognition &
Recognize static visual information from a single frame or a set of frames without requiring correct temporal order, including object/entity recognition, attribute recognition, scene understanding, and OCR-based text reading. \\
L1 & Frame Only & Basic Counting &
Count the number of target objects/entities appearing in the video (static counting). Aggregation across frames is allowed, but the correct answer should not depend on frame ordering. \\
L1 & Frame Only & Numerical Calculation &
Read numerical information from the visual stream (e.g., house numbers, clocks/timestamps shown on screen, displayed formulas) and perform simple arithmetic/comparisons. The task should not involve multi-hop reasoning beyond straightforward computation. \\
\midrule
L1 & Frames \& Audio & Cross-Modal Semantic Consistency &
Judge whether visual content and audio are semantically/affectively consistent (e.g., whether the perceived mood, event type, or semantic content matches). The question should not rely on temporal-change reasoning that belongs to Level~2. \\
L1 & Frames \& Audio & Audio-Guided Visual Description &
Use an audio cue to temporally localize a moment, then answer what is shown visually at that moment. Distractors should be time-near and/or visually similar to enforce precise alignment. \\
L1 & Frames \& Audio & Vision-Guided Audio Description &
Use a visual cue to temporally localize a moment, then answer what is said/heard (speech content, sound event, or audio attribute) at that moment. \\
L1 & Frames \& Audio & Visual-Audio Collaborative Reasoning &
Require genuine audio-visual \emph{collaboration} through temporal alignment: neither the video frames alone nor the audio alone should be sufficient to reliably answer; the correct answer depends on associating what is heard with what is seen at the corresponding time. \\
\midrule
L2 & Action \& Motion & Fine-Grained Action Recognition &
Recognize fine-grained actions and discriminate between visually similar actions using motion dynamics over time (not solvable from a single still frame). Examples include specific sports moves or detailed procedural actions. \\
L2 & Action \& Motion & Repetitive Action Counting &
Count repetitions of a recurring action where correct temporal order is essential (shuffling frames should break the ability to count). Actions should not be too fast to be missed under low sampling rates. \\
L2 & Action \& Motion & Temporal Action Localization &
Localize when an action happens (start/end or event-relative timing such as “right after X”). Exact timestamps are not required; event-based references are preferred. \\
L2 & Action \& Motion & Motion Trajectory Formation &
Infer the motion trajectory/path of an object or person (e.g., straight vs. curved vs. detouring), requiring aggregation over time. \\
L2 & Action \& Motion & Motion Properties Analysis &
Analyze motion properties over time, including direction (left/right/up/down), speed changes, amplitude changes, and frequency changes. \\
\midrule
L2 & Order & Object Appearance Ordering &
Determine the temporal order of object/person appearances. The question must rely on the timeline such that frame shuffling harms solvability. \\
L2 & Order & Event Sequence Ordering &
Determine the temporal order of multiple events/steps (often global ordering). The correct answer must require understanding the sequence rather than recognizing a single moment. \\
\midrule
L2 & Change & Temporal Periodicity Detection &
Detect whether actions/events exhibit periodicity or cycles and characterize the cycle structure. Correct ordering is crucial for identifying repetition patterns. \\
L2 & Change & Entity Existence Change Detection &
Track changes in entity existence (appear/disappear/leave the scene). Questions should explicitly incorporate both before/after states to reflect change over time. \\
L2 & Change & Entity Attribute Change Detection &
Track changes in entity attributes or physical state over time (e.g., color change, melting/freezing). The question must require comparing states across different time points. \\
L2 & Change & Scene Transformation Detection &
Track scene transformations such as camera cuts, location shifts (indoor$\rightarrow$outdoor), or background transitions, requiring temporal understanding. \\
\midrule
L2 & Temporal Reasoning & Causal Reasoning &
Infer cause-effect relations grounded in the video timeline (cause precedes effect). Both cause and effect should be supported by video evidence (preferably visual), not by external commonsense alone. \\
L2 & Temporal Reasoning & Future Event Prediction &
Predict a likely future event based on observed dynamics/trends in the video. The task must be temporally grounded (not solvable by reading a single chart frame). \\
\midrule
L3 & Complex Plot Comprehension & Narrative Turning Point Detection &
Identify key turning points or plot reversals and select the best explanation grounded in preceding narrative buildup. \\
L3 & Complex Plot Comprehension & Narrative Clues Inference &
Infer hidden truths or unstated implications by integrating multiple narrative clues across time (visual details, behaviors, dialogue context). Answers should not be directly readable from any single moment. \\
L3 & Complex Plot Comprehension & Symbolic/Metaphorical Interpretation &
Interpret symbolism or metaphor (objects/events representing deeper meanings) in context, requiring narrative-level understanding. \\
L3 & Complex Plot Comprehension & High-Order Narrative Deconstruction &
Explain higher-order storytelling/cinematic devices (e.g., montage, multi-thread narrative, stylistic camera language) and why they create certain effects (e.g., humor, suspense), grounded in the video. \\
\midrule
L3 & Video-Based Knowledge Acquisition & Professional Knowledge Acquisition &
Acquire domain knowledge from lectures/tutorials and answer questions that require attending to the specific content presented (not solvable by generic background knowledge alone). \\
L3 & Video-Based Knowledge Acquisition & General Skills Acquisition &
Acquire everyday procedural skills from instructional videos (e.g., cooking/repair) and answer questions about correct steps, conditions, or error checking, grounded in the demonstrated procedure. \\
\midrule
L3 & Social Behavior Analysis & Individual Social Cognition &
Infer an individual’s intent, emotion, attitude, or mental state using multimodal evidence (behavior, facial expressions, tone, context) across the video. \\
L3 & Social Behavior Analysis & Dyadic Interaction Dynamics &
Analyze interpersonal dynamics between two people (relationship evolution, status differences, conversational tone) by integrating evidence across multiple exchanges. \\
L3 & Social Behavior Analysis & Collective Dynamics Analysis &
Analyze group-level dynamics (emergent roles, alliances, cooperation vs. conflict, conversational dominance) requiring integration across multiple participants and time spans. \\
\midrule
L3 & Physical World Reasoning & Entity Persistence Tracking &
Reason about object persistence under occlusion or temporary absence, inferring continued existence and plausible location consistent with physical continuity. \\
L3 & Physical World Reasoning & Spatial Understanding &
Reason about the 3D world depicted by the video (relative positions, depth/front-back relations, routes/directions), rather than simply describing 2D image layout. \\
L3 & Physical World Reasoning & Counterfactual Reasoning &
Perform grounded counterfactual reasoning: predict outcomes under altered conditions while respecting the physical rules implied by the video context. \\
L3 & Physical World Reasoning & Counterintuitive Comprehension &
Detect and explain phenomena that violate physical intuition or laws (e.g., magic tricks, VFX), identifying what is unusual and why it contradicts everyday physics. \\

\end{longtable}
\endgroup

\noindent\begin{minipage}{\linewidth}
    \centering
    \includegraphics[width=\linewidth]{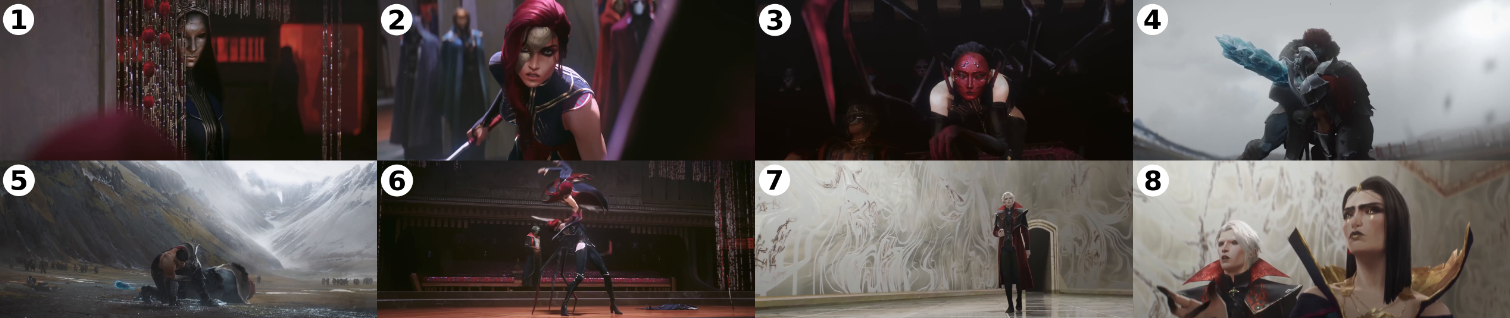}
    \captionof{figure}{\textbf{An example in Level 1:} Visual Recognition}
    \label{fig:level1_case}

    \raggedright
    \vspace{0.4em}

    <Question 1>: In the ballroom scene, what color cloak was the assassin wearing at the beginning?\\[0.3em]
    <Options 1>: A. Brown. B. Grey. C. White. D. Red. E. Purple. F. Black. G. Green. \textcolor{red}{H. Blue.}

    \medskip
    <Question 2>: In the ballroom scene, what animal is the monster fighting the assassin based on?\\[0.3em]
    <Options 2>: \textcolor{red}{A. Spider.} B. Jaguar. C. Butterfly. D. Hornet. E. Falcon. F. Praying Mantis. G. Lion. H. Fly.

    \medskip
    <Question 3>: What color are the white-haired man's eyes?\\[0.3em]
    <Options 3>: A. Brown. B. Green. C. Black. D. Purple. E. Yellow. F. Blue. G. White. \textcolor{red}{H. Red.}

    \medskip
    <Question 4>: In the ice field scene, what weapons are used by the two opposing sides?\\[0.3em]
    <Options 4>: \textcolor{red}{A. Club and Axe.} B. Both sides use axes. C. Both sides use staves. D. Longsword and Axe. E. Dagger and Axe. F. Dagger and Longsword. G. Staff and Dagger. H. Both sides use daggers.
\end{minipage}

\noindent\begin{minipage}{\linewidth}
    \centering
    \includegraphics[width=\linewidth]{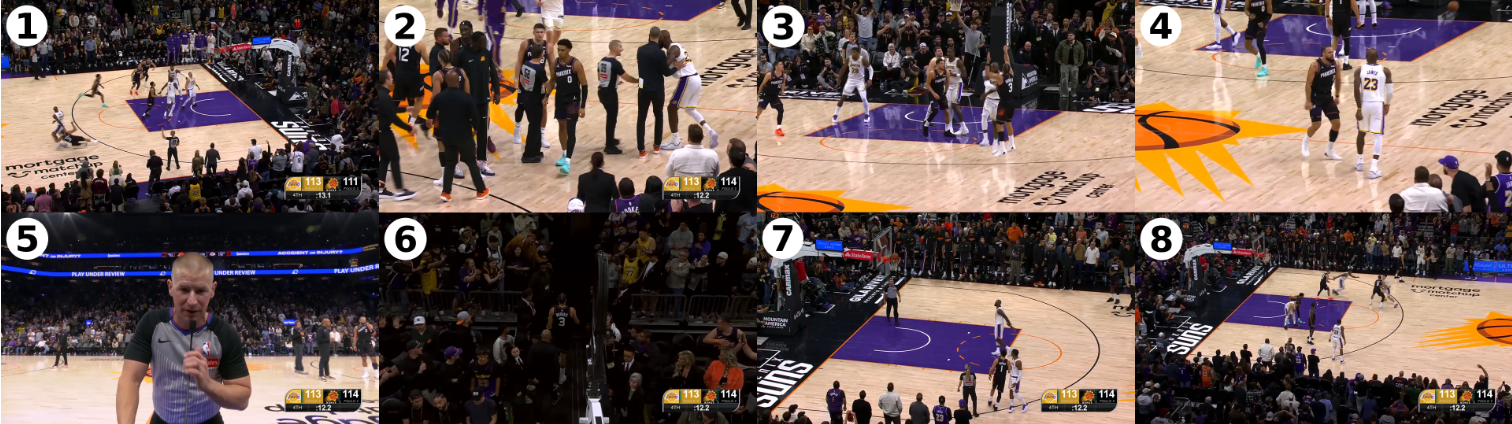}
    \captionof{figure}{\textbf{An example in Level 2:} Temporal Reasoning}
    \label{fig:level2_case}

    \raggedright
    \vspace{0.4em}

    <Question 1>: Why did the Suns' player \#3 leave the court when the score was 113:114?\\[0.3em]
    <Options 1>: A. Because he could not continue due to excessive physical exhaustion. B. Because he could not continue due to a rib injury from a collision. C. Because he could not continue due to an ankle injury. D. Because he was protesting the officiating by refusing to play. \textcolor{red}{E. Because he was ejected due to a technical foul.} F. Because it was time for a rotation substitution. G. Because he was ejected due to regular fouls. H. Because he was subbed out by the coach due to poor performance.

    \medskip
    <Question 2>: Why did the Lakers' player \#23 go to the free-throw line when the score was 113:114?\\[0.3em]
    <Options 2>: A. Because the Suns' center committed a defensive three-second violation. B. Because a Suns player committed a blocking foul while defending the Lakers' player \#23's layup. C. Because the Suns' coach argued with the referee and was assessed a technical foul. \textcolor{red}{D. Because a Suns player committed a technical foul, and the Lakers' player \#23 chose to execute the free throw himself.} E. Because the Suns requested a timeout when they had none remaining. F. Because the Suns had more than 5 players on the court after a timeout, and the Lakers' player \#23 chose to execute the free throw himself. G. Because the Suns had fewer than 5 players on the court after a timeout, and the Lakers' player \#23 chose to execute the free throw himself. H. Because the Lakers were fouled tactically on a fast break with no defender in front, and the Lakers' player \#23 chose to execute the free throw himself.
    
    \medskip
    <Question 3>: Why did the Suns' player \#3 feel dissatisfied and angry when the score was 113:114?\\[0.3em]
    <Options 3>: A. Fans on the sidelines were booing him. B. He believed the referee's officiating standards were inconsistent and biased. C. He believed the team did not call a timeout in time to make adjustments. D. He felt dissatisfied with the coach's substitution adjustments. \textcolor{red}{E. A Lakers player made a motion to crash into him.} F. He repeatedly received "grenade" passes from teammates and had a difficult time on the court. G. He believed the referee made a serious missed call. H. He missed a crucial three-point shot.

    \medskip
    <Question 4>: What impact did missing that technical free throw have on the tactics of both sides for the final possession?\\[0.3em]
    <Options 4>: A. Neither the offensive nor defensive strategies are affected, as the game has entered garbage time. B. The offensive team can choose to play conservatively to force overtime, or take some risk to attack actively; the defensive team's strategy remains unchanged. \textcolor{red}{C. The offensive team needs to actively complete the last attack; the defensive team's strategy remains unchanged.} D. The offensive team can only choose to shoot a three-pointer on the final possession; the defensive team only needs to focus on defending beyond the three-point line. E. The offensive team needs to complete the attack quickly to fight for an extra offensive possession; the defensive team's strategy remains unchanged. F. The offensive team's strategy remains unchanged; the defensive team only needs to focus on defending the paint. G. The offensive team needs to make a three-pointer to overcome the deficit; the defensive team's strategy remains unchanged. H. The offensive team's strategy remains unchanged, needing only to run out the clock to win; the defensive team needs to foul quickly to gain an extra offensive possession.
\end{minipage}

\noindent\begin{minipage}{\linewidth}
    \centering
    \includegraphics[width=\linewidth]{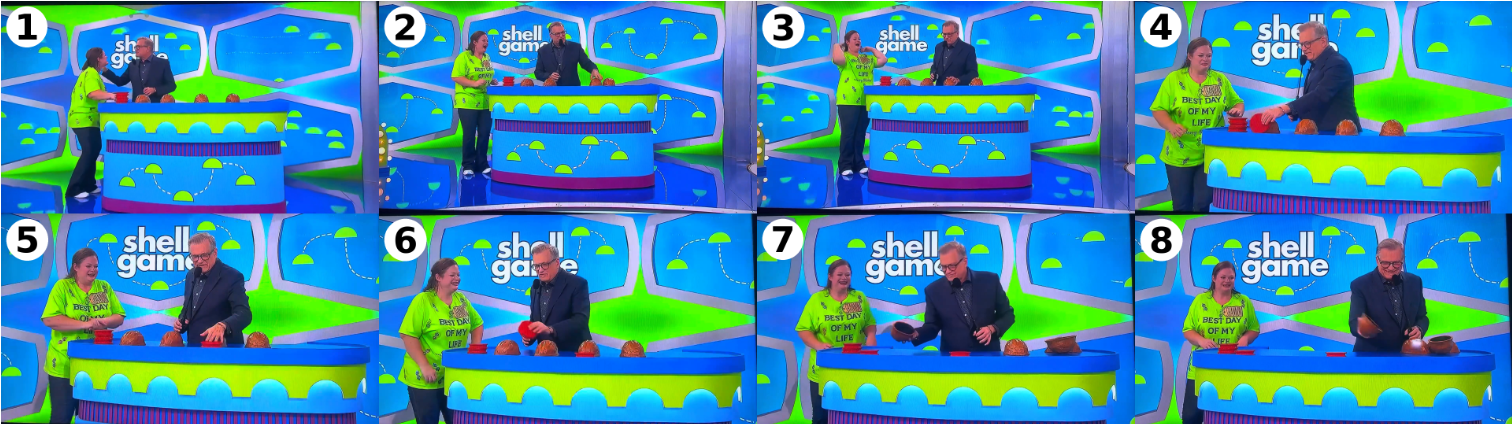}
    \captionof{figure}{\textbf{An example in Level 3:} Entity Persistence Tracking}
    \label{fig:level3_case}

    \raggedright
    \vspace{0.4em}

    <Question 1>: Does the ball exist underneath any of the shells?\\[0.3em]
    <Options 1>: A. No. \textcolor{red}{B. Yes.} C. Cannot be determined.

    \medskip
    <Question 2>: Underneath which shell is the ball located at the end?\\[0.3em]
    <Options 2>: A. There is no ball under any shell. B. The third shell. C. The sixth shell. D. The second shell. E. The seventh shell. F. The fifth shell. \textcolor{red}{G. The fourth shell.} H. The first shell.

    \medskip
    <Question 3>: The host performed a total of two shell swaps (defining a single swap as an instance where all shells return to an approximately straight line). Underneath which shell was the ball located after the first swap?\\[0.3em]
    <Options 3>: A. There is no ball under any shell. B. The seventh shell. \textcolor{red}{C. The fourth shell.} D. The fifth shell. E. The sixth shell. F. The second shell. G. The third shell. H. The first shell.

    \medskip
    <Question 4>: The host performed a total of two shell swaps (defining a single swap as an instance where all shells return to an approximately straight line). Underneath which shell was the ball located initially?\\[0.3em]
    <Options 4>: A. The seventh shell. B. The fourth shell. C. The fifth shell. \textcolor{red}{D. The third shell.} E. The second shell. F. There is no ball under any shell. G. The first shell. H. The sixth shell.
\end{minipage}

\end{document}